\documentclass{article}

\PassOptionsToPackage{numbers, sort&compress}{natbib}

\usepackage[preprint]{neurips_2024}
\usepackage[ruled,linesnumbered]{algorithm2e}
\usepackage{subcaption}

\usepackage[utf8]{inputenc} %
\usepackage[T1]{fontenc}    %
\usepackage{url}            %
\usepackage{booktabs}       %
\usepackage{amsfonts}       %
\usepackage{nicefrac}       %
\usepackage{microtype}      %
\usepackage{xcolor}         %

\usepackage{graphicx}
\usepackage{booktabs}

\usepackage[accsupp]{axessibility}  %

\usepackage{xspace}
\usepackage{mathtools}
\usepackage{enumitem}
\usepackage{wrapfig}
\definecolor{citecolor}{HTML}{2779af}
\definecolor{linkcolor}{HTML}{c0392b}
\usepackage[pagebackref=false,breaklinks=true,colorlinks,bookmarks=false,citecolor=citecolor,linkcolor=linkcolor]{hyperref}

\newcommand{\fullmethod}{Reinforcement Learning via Auxiliary Task Distillation\xspace}
\newcommand{\method}{AuxDistill\xspace}
\newcommand{\nodist}{AuxDistill (No-Distill)\xspace}
\newcommand{\curric}{RL Curriculum\xspace}
\newcommand{\pick}{Category Pick\xspace}
\usepackage[textsize=tiny]{todonotes}
\newcommand{\as}[1]{\todo[color=red!20, size=\tiny]{AS: #1}}

\renewcommand{\as}[1]{}

\DeclarePairedDelimiterX{\infdivx}[2]{(}{)}{%
  #1\;\delimsize\|\;#2%
}
\newcommand{\kl}{D_{\text{KL}}\infdivx}

\usepackage{multirow,tabularx}
\usepackage{colortbl}
\definecolor{Gray}{gray}{0.5}
\definecolor{NewBlue}{rgb}{0.95, 0.95, 1.0}

\usepackage{cleveref}

\title{Reinforcement Learning via Auxiliary Task Distillation}

\author{%
  Abhinav Narayan Harish$^{1}$,
  Larry Heck$^{2}$,
  Josiah P. Hanna$^{1}$,
  Zsolt Kira$^{2}$,
  Andrew Szot$^{2}$\\
  $^{1}$University of Wisconsin -- Madison , $^{2}$Georgia Tech
}

\begin{document}

\maketitle 
\begin{abstract}
    We present Reinforcement Learning via Auxiliary Task Distillation (AuxDistill), a new method that enables reinforcement learning (RL) to perform long-horizon robot control problems by distilling behaviors from auxiliary RL tasks. AuxDistill achieves this by concurrently carrying out multi-task RL with auxiliary tasks, which are easier to learn and relevant to the main task. A weighted distillation loss transfers behaviors from these auxiliary tasks to solve the main task. We demonstrate that AuxDistill can learn a pixels-to-actions policy for a challenging multi-stage embodied object rearrangement task from the environment reward without demonstrations, a learning curriculum, or pre-trained skills. AuxDistill achieves $2.3 \times$ higher success than the previous state-of-the-art baseline in the Habitat Object Rearrangement benchmark and outperforms methods that use pre-trained skills and expert demonstrations.

\end{abstract}

\section{Introduction}

While reinforcement learning (RL) is successful in a variety of settings including games~\cite{mnih2013playing,berner2019dota,silver2017mastering}, chatbots~\cite{shah2016interactive, liu2017end,nakano2021webgpt,achiam2023gpt,touvron2023llama}
, and robotics~\cite{akkaya2019solving,qi2023hand,qt-opt}, long-horizon tasks such as embodied object rearrangement where an embodied agent must rearrange objects to target positions still remains a challenge~\cite{batra2020rearrangement}. 
Object rearrangement requires learning heterogeneous behaviors like picking, navigating, placing, and opening concurrently with sequencing these behaviors to solve the overall task all while operating from egocentric vision.
Furthermore, the behaviors are interdependent, meaning the robot can only learn how to pick an object if it has first learned how to open the fridge containing the object.
Since object rearrangement requires interacting with objects across house-scale spaces with low-level control, the episodes consist of thousands of low-level control steps, exacerbating the problem of credit assignment~\cite{harutyunyan2019hindsight,ni2024transformers}. 
Overall, the problems of concurrently learning low-level control, high-level decision-making, interdependent task stages, and episodes with many time steps make object rearrangement challenging even with dense rewards. 
While some prior methods have applied end-to-end RL to rearrangement problems, they require excessive experience in simplified simulators~\cite{berges2023galactic} or expert skill demonstrations~\cite{huang2023skill}. 
Other prior work has made such complex tasks more tractable by decomposing the full task into a hierarchy of separately trained skills that a high-level policy sequences together to solve the overall task. 
However, such hierarchical methods suffer from compounding errors between skills~\cite{gu2022multi} and dynamically selecting between skills. 

We therefore present \textbf{\fullmethod (\method)}, a method for training policies for long-horizon tasks from scratch with end-to-end RL from reward alone.
\method also learns in auxiliary tasks using a multi-task RL framework and transfers the knowledge from these auxiliary tasks to help solve the desired ``main'' task of interest.
Importantly, \method concurrently learns the main task and all auxiliary tasks, which are all initialized from scratch and not pre-trained. These auxiliary tasks help because they are \emph{easier to learn} with RL than the main task and contain \emph{relevant behaviors} for the full task. 
For example, in object rearrangement, the agent should also practice picking up objects in isolation from the complexities of the full task.
Unlike curriculum learning in RL~\cite{narvekar2020curriculum,dennis2020emergent,azad2023clutr,fang2022active}, which also leverages easier tasks to learn harder tasks, \method does not have distinct curriculum phases. Instead, it has a single training objective for distilling sub-behaviors into the overall task behavior.
Crucially, the agent end-to-end learns which auxiliary behaviors are relevant for the main task. This relevance is encoded by a scalar weight value determined by the relevance of the robot state to the auxiliary task and it grounded in the oracle task plan. 
A distillation objective transfers behaviors from relevant auxiliary tasks by encouraging the policy to act consistently between the main and auxiliary tasks for relevant states in the main task.
For instance, in object rearrangement, when the robot is near a drawer with the target object inside, it is supervised with the distillation loss from an ``open drawer" auxiliary task.
Despite the policy also concurrently learning the ``open drawer'' auxiliary task from scratch, \method learns the auxiliary tasks faster since they are easier. 
Thus, the distillation loss is a useful dense per-time step supervision signal that helps overcome the challenges of learning from the main task reward alone.

We empirically demonstrate that \method outperforms a variety of baselines in terms of success rate on the home rearrangement benchmark in Habitat 2.0~\cite{szot2021habitat}. These include baselines such as  hierarchical RL, end-to-end RL with and without a curriculum, and imitation learning.
On rearrangement episodes where objects start in open receptacles, \method achieves $ 1.4 \times $ higher success than baselines and $2.6 \times$ higher success in harder episodes where objects can start in closed receptacles.
These results highlight the value of \method to be able to learn the entire rearrangement task end-to-end with RL.
We show results beyond object rearrangement on a category-conditioned object manipulation task where \method achieves $  1.75 \times$ higher success than baselines.
We also conduct extensive ablation studies where we analyze the properties of auxiliary tasks needed by \method, and find that \method is generally robust to the choice of auxiliary tasks.
An analysis of the distillation objective reveals the importance of the distillation loss, with the absence of the distillation loss leading to no success on the rearrangement task. 
All code is available at \url{https://github.com/absdnd/aux_distill}

\vspace{-10pt}
\section{Related Work}

Some prior works address complex and long-horizon decision-making problems using a hierarchical breakdown of skills. One such category of approaches is option-critic methods~\cite{sutton1999between,bacon2017option,zhang2019dac}, which seek to learn options to temporally abstract the high-level policy decision making. However, discovering and learning such options is unstable and results in a challenging credit assignment problem. Another line of work first trains or is given pre-trained skills and then learns a high-level policy to sequence these skills together to complete longer tasks~\cite{xia2020relmogen,karkus2020beyond,dalal2021accelerating,hafner2022deep}. Using such a hierarchy results in compounding errors resulting from sequencing skills that were not trained to properly transition between each other~\cite{vezzani2022skills,chen2023sequential,mishra2023generative,lee2021adversarial,gu2022multi}. Our approach does not utilize a hierarchical policy, and can solve long horizon tasks better than hierarchical methods by utilizing a single policy and thus avoiding compounding errors that emerge by sequencing independently trained policies.  

Like our method, prior works have tackled model-free end-to-end RL to solve long, complex tasks.
Scaling PPO~\cite{schulman2017proximal} training has enabled robots to learn complex locomotion behaviors~\cite{rudin2022learning,agarwal2023legged,fu2023deep,kumar2021rma,radosavovic2023learning,katara2023gen2sim}, manipulate unseen objects in a robotic hand~\cite{qi2023hand}, and play competitive video games~\cite{berner2019dota}.
Likewise, \citet{berges2023galactic} showed scaling end-to-end RL can learn object rearrangement. However, their approach requires training agents for billions of environment steps with a simplified kinematic physics simulation. 
We show results in the full dynamic Habitat simulation, achieving better results in an order of magnitude fewer environment interactions, and show results in harder rearrangement tasks. 
Other works have incorporated auxiliary objectives in RL to boost performance in embodied tasks ~\cite{ye2021auxiliary,ye2021auxiliary2,gregor2019shaping, kuo2023structure}.
These works add auxiliary self-supervised prediction objectives, whereas our method adds an auxiliary policy learning task via multi-task RL.
 \citet{jia2022improving} also uses easier auxiliary RL tasks that specialist agents learn to complete and then distill back into a generalist agent. 
While our work also uses RL in easier tasks to boost performance, it doesn't require alternating between learning specialist and generalist policies. 

\looseness=-1 Works have also attempted to learn a curriculum to leverage easier tasks to learn harder tasks that require composing multiple behaviors. Some works learn curriculum generation policies that adjust the environment to suit the agent's capabilities~\cite{dennis2020emergent,azad2023clutr,fang2022active}. In other works, a curriculum naturally emerges as a result of downstream training \cite{baker2019emergent}. Finally, some works hand design curricula that change aspects of the environment, such as the starting and goal distributions depending on the agent performance during training~\cite{rudin2022learning}. Our work shares a similar spirit of leveraging easier auxiliary tasks to learn a harder task but it does not enforce an explicit curriculum and instead learns all tasks simultaneously.

\begin{figure*}[t]
  \centering
  \includegraphics[width=\textwidth]{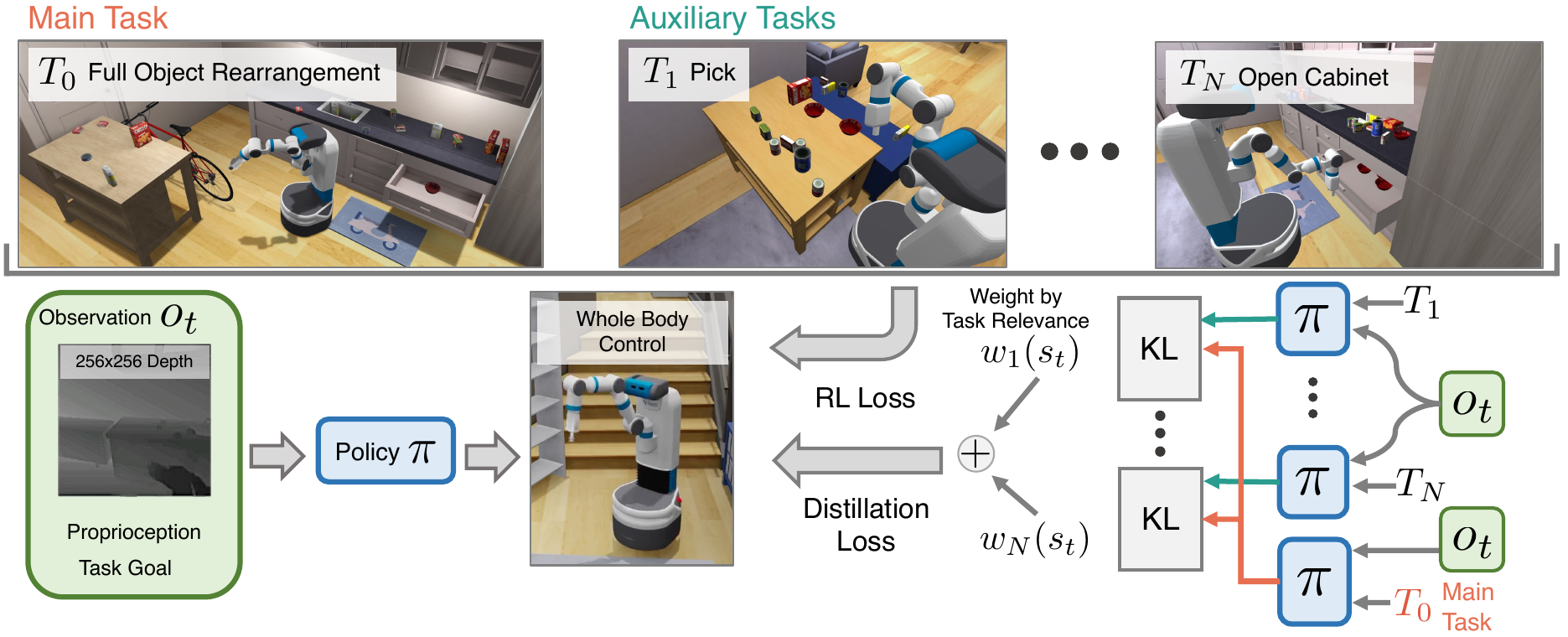}
  \caption{
    \method learns a rearrangement policy operating from egocentric depth perception and coordinate-based task specification. The full object rearrangement task decomposes into modular abilities that can be learned by auxiliary  task with indicator vectors ${T}_{1} \cdots {T}_{N}$ which are trained along with the main task using end-to-end RL. 
    The distillation loss is computed as a weighted combination of the task relevance of $ o_t$ in the main task $ T_0$ under all auxiliary tasks. 
    The task relevance function computes $ w_i(s_t)$ based on the relevance of the current observation and robot state to the auxiliary task $T_{i}$. 
    The distillation loss and RL-training loss are then used to update the policy.
    }
  \label{fig:method}
\end{figure*}

\section{Method}
\label{sec:method} 

We present \fullmethod (\method), a new method for training long-horizon policies from scratch using rewards alone. 
Learning from reward alone in complex tasks like embodied object rearrangement is challenging because an agent needs to combine thousands of low-level actions controlling the arm and base, operate from egocentric visual perception, and dynamically sequence distinct behaviors such as navigating, picking, placing, and opening. 
\method addresses these challenges of RL in complex problems, like embodied rearrangement, by learning to leverage knowledge from easier auxiliary tasks related to the desired task we are trying to solve. Unlike a curriculum, which learns gradually harder tasks in stages, \method learns the desired task concurrently with the auxiliary task and includes a novel distillation mechanism to transfer knowledge from easier to harder tasks. 

\subsection{Preliminaries}
\label{sec:prelims}
Our problem is formulated as a goal-specified Partially-Observable Markov Decision Process (POMDP) defined as a tuple $ \mathcal{M} = \left( \mathcal{S}, \mathcal{O}, \mathcal{A}, \mathcal{P}, \mathcal{R}, \mathcal{G}, \rho, \gamma \right) $ with underlying state space $ \mathcal{S}$, observation space $ \mathcal{O}$, action space $ \mathcal{A}$, transition function $\mathcal{P}$, reward function $ \mathcal{R}$, goal space $ \mathcal{G}$, initial state distribution $ \rho$ and discount factor $ \gamma$.   For a task like rearrangement, the goal space $\mathcal{G}$ is specified using the 3D coordinate of the object's start location and the goal location where the object has to be placed. Our objective is to learn a goal-conditioned policy $ \pi(a \mid o, g)$ mapping an observation $ o $ and goal $ g$  to an action $ a$ that maximizes the sum of discounted rewards $ \mathbb{E}_{s_0 \sim \rho_0, g \sim \mathcal{G}} \sum_{t} \gamma^{t} \mathcal{R}(s_t, g)$. 

\subsection{\fullmethod}

 The core idea of \method is to learn a policy in a difficult desired ``main'' task by transferring knowledge from easier auxiliary tasks. This is done without using an explicit curriculum by learning the main and auxiliary tasks in a single training loop. We refer to the main task we wish to learn as $ \mathcal{M}_0$ with a task identifier  $T_{0}$.
 We define $N$ \emph{auxiliary tasks}, defined as $ \left\{ \mathcal{M}_i \right\}_{n=1}^{N}$ with task identifiers $ \left\{ {T}_i \right\}_{n=1}^{N}$. The auxiliary tasks $ \left\{ \mathcal{M}_i \right\} $ share the same state, observation, goal, and action space as the target task $ \mathcal{M}_0$. Each $ \mathcal{M}_i$ has a separately defined starting state distribution and reward function. We assume these auxiliary tasks are related to the full task yet are easier to solve than the full task. Specifically, the auxiliary tasks can be easier instantiations or sub-parts of the overall task. 
For example, in rearrangement, we define the auxiliary tasks in terms of interactions the agent needs to complete the entire rearrangement episode, such as picking, placing, and opening.
Prior works use similar task definitions to train skills for hierarchical policies in rearrangement~\cite{gu2022multi}, but these works suffer from a two-stage pipeline of first needing to separately train each skill and then decide how to combine them.
\method directly learns the full task from scratch while also performing better.

\method learns a single policy with RL that concurrently learns to perform the main and auxiliary tasks. We illustrate the implementation of \method in ~\Cref{fig:method}.
This policy, parameterized by $ \theta$, is expressed as $ \pi_\theta (a_t \mid o_t , g, T) $ for observation $ o_t$, episode goal  $ g$, and per-$ \mathcal{M}$ task identifier $ T$ which is encoded as a one-hot embedding in a vector which has the same size as the maximum number of auxiliary tasks. Note that since all tasks share the same observation and action space, $ \pi_\theta$ can act and observe in all tasks based on the input task identifier $T$. 

\method updates the policy based on an average RL loss from the main task and all auxiliary tasks. 
Let $\mathcal{L}_{\mathcal{M}_i}^{\text{RL}}(\theta)$ denote the RL loss  for $ \pi_\theta$ in MDP $ \mathcal{M}_i$. These auxiliary tasks are designed to capture a subset of the main task which are easier to accomplish than the main task. 
To compute these losses, we assume we can collect experience in the auxiliary tasks.
For example, to collect experience in a ``pick object'' auxiliary task in object rearrangement, the robot is spawned close to the object.
We then update based on the average of the task and auxiliary task losses: $ \frac{1}{N} \sum_{i=0}^{N}$ $\mathcal{L}_{\mathcal{M}_i}^{\text{RL}}(\theta)$. 
Intuitively, $ \pi_\theta $ will first learn to complete the easier auxiliary tasks, and this auxiliary task competency can aid in solving the main task.
This shares a similar insight as curriculum learning, except all tasks are learned concurrently, and the curriculum stages are naturally induced by the policy naturally learning on easier tasks first. 

In addition to optimizing the average RL loss between the main and auxiliary tasks, \method also optimizes a \textit{distillation loss} that encourages the policy to transfer relevant knowledge from the auxiliary tasks to the main task.
To achieve this, for a particular observation $o_t $, we consider the policy distribution $\pi_\theta ( \cdot \mid o_t, g, T) $ under different task identifiers $T$.
For compactness, notate $ \pi_\theta( \cdot \mid o_t, g, T_i)$ where $ T_i$ indicates the task identifier for $ \mathcal{M}_i$ as $ \pi_\theta^{T_i} \left( o_t \right) $.
We want $ \pi_\theta^{T_0} $ (the policy in the main task) to match the behaviors from the policy in the relevant auxiliary tasks $ \pi_\theta^{T_i} $.
We measure this relevance of an auxiliary task $\mathcal{M}_i $ to the main task at time step $t$ via an \textit{auxiliary task relevance function} $w_i(s_t)$. 
This function denotes how much the knowledge from $ \mathcal{M}_i$  should apply to $ \mathcal{M}_0$ in the underlying simulator state $ s_t$. 
This relevance function can be grounded in the task plan generated by an oracle planner. For example, consider the state of the robot before picking up the object. In this state, the \textit{pick} auxiliary task would be relevant to the main task, and the \textit{place} auxiliary task would not. If an episode requires opening a fridge before accessing the object, the relevant task before the fridge is opened would be \textit{open-fridge}. Note that computing the distillation weight can utilize oracle knowledge of the simulator state (for instance, whether the object is inside a fridge or a cabinet) since this information is only provided as a training signal and not used during inference. This information is utilized by all methods we compare against, including our strongest baseline~\cite{huang2023skill}. 

We can then distill experience from $ \mathcal{M}_i$ by supervising the policy to match the action distribution of the target task $\mathcal{M}_{0}$. The distillation loss is computed as the KL-divergence of the action distribution under $ \mathcal{M}_i$ and $\mathcal{M}_{0}$ weighted by the relevance of auxiliary task $\mathcal{M}_{i}$ given by  $w_i(s_t)$. 
The overall loss function of \method, including the distillation loss, is then:
\begin{align}
    \label{eq:main_loss}
  \frac{1}{N} \sum_{i=0}^{N} \mathcal{L}_{\mathcal{M}_i}^{\text{RL}}(\theta) + \lambda \sum_{i=1}^{N} w_{i}(s_t) \kl{\pi_\theta^{T_0} (o_t)}{\pi_\theta^{T_i} (o_t)}
\end{align} 
Here $\lambda$ represents the distillation weight relative to other PPO losses. Note that our formulation relies on the \textit{auxiliary} tasks being easier to solve and relevant to the main task. Distillation using the task relevance function offers an additional supervisory signal that the policy can optimize along with its own reward. This doesn't impose the large overhead of hierarchical RL methods where each auxiliary task must be pre-trained in isolation with its own reward function and goal specification. Auxiliary tasks are flexibly incorporated to address parts of the task that are challenging for the agent to learn.

\subsection{Implementation Details}
\label{sec:impl-details}

We use PPO~\cite{schulman2017proximal} to compute the RL loss $ \mathcal{L}_{\mathcal{M}_i}^{\text{RL}}$ in \Cref{eq:main_loss}. 
We create $ M$ environment instances for each of the main and $ N$ auxiliary tasks and vectorize them for parallel action execution.
We rollout the policy in all these $ M(N+1)$ environments and collect a batch of data.
We then compute \Cref{eq:main_loss}, using the PPO loss, update the policy, and then repeat this process.

The policy is represented as a 2-layer LSTM network with 512 hidden units per layer, similar to the architecture employed in ~\cite{szot2021habitat}. 
In our experiments, $ o_t$ is a depth or RGB egocentric image, which we encode with a ResNet50~\cite{he2016deep} network. The goal and the proprioceptive state of the agent are concatenated with the visual embeddings along with the index of the POMDP $ \mathcal{M}_i$, for which we use a one-hot embedding. This vector is then passed as an input to the LSTM. A linear projection on the output of the LSTM then produces an action to execute on the next step in the environment and a vector of value-predictions corresponding to each of the tasks $\mathcal{M}_{i}$ used during training.

Our policy is learned via a multi-task RL formulation, with the main and auxiliary tasks being concurrently learned. Each of the tasks may have different reward magnitudes and relative performance during training. To address this, we use PopArt~\cite{hessel2019multi}, with $ \beta = 3e^{-4}$ to enable learning across different return scales. Additional \method details are in \Cref{sec:method-details} of the supplementary.

\begin{table*}[t]
  \centering
  \resizebox{0.98 \textwidth}{!}{
    \scalebox{0.6}{
\begin{tabular}{lcccccc}
\toprule
& \multicolumn{3}{c}{\textbf{Train} (Seen)} & \multicolumn{3}{c}{\textbf{Eval} (Unseen)} \\
\cmidrule(rl){2-4} \cmidrule(rl){5-7}
Method & \textbf{All Episodes} & \textbf{Easy}  &  \textbf{Hard}  & \textbf{All Episodes}  &  \textbf{Easy}  &  \textbf{Hard}  \\
\midrule
\color{Gray} \textbf{M3 (Oracle)}~\cite{gu2022multi} & \color{Gray}  27 {\scriptsize $ \pm $ 0} & \color{Gray}  57 {\scriptsize $ \pm $ 2}  & \color{Gray}  12 {\scriptsize $ \pm $ 1}  & \color{Gray}  28 {\scriptsize $ \pm $ 0}  & \color{Gray}  58 {\scriptsize $ \pm $ 2} & \color{Gray} 13 {\scriptsize $ \pm$ 2} \\
\midrule
\textbf{M3}~\cite{gu2022multi} &  25 {\scriptsize $\pm$ 1}  & {56} {\scriptsize $ \pm$ 1} & 9 {\scriptsize $\pm$ 2} &  13 {\scriptsize $ \pm$ 1} & 53 {\scriptsize $ \pm$ 4} & 0 {\scriptsize $ \pm $ 0} \\
\textbf{Galactic}~\cite{berges2023galactic} &  - & {37} {\scriptsize $ \pm$ 0} & -  & -  & 26 {\scriptsize $ \pm$ 0} & - \\
\textbf{Monolithic RL} &  0 {\scriptsize $\pm$ 0}  &  0 {\scriptsize $\pm$ 0}  &  0 {\scriptsize $\pm$ 0} &  0 {\scriptsize $\pm$ 0} &  0 {\scriptsize $\pm$ 0} &  0 {\scriptsize $\pm$ 0} \\
\textbf{Skill Transformer}~\cite{huang2023skill} &  25 {\scriptsize $ \pm$ 1}   &  44{\scriptsize $\pm$ 1} &  16 {\scriptsize $ \pm$ 1} &  23 {\scriptsize $ \pm$ 1}  & 37{\scriptsize $\pm$ 1} & 16{\scriptsize $\pm$ 1} \\
\textbf{\method (No Distillation)} & 0{\scriptsize $\pm$ 0} & 0{\scriptsize $\pm$ 0} & 0{\scriptsize $\pm$ 0} & 0{\scriptsize $\pm$ 0} & 0{\scriptsize $\pm$ 0} & 0{\scriptsize $\pm$ 0} \\ \textbf{RL Curriculum} & 0{\scriptsize $\pm$ 0} & 1{\scriptsize $\pm$ 0} & 0{\scriptsize $\pm$ 0} & 0{\scriptsize $\pm$ 0} & 1{\scriptsize $\pm$ 1} & 0{\scriptsize $\pm$ 0} \\ 
\textbf{\method} & \textbf{49}{\scriptsize $\pm$ 2} & \textbf{74}{\scriptsize $\pm$ 2} & \textbf{36}{\scriptsize $\pm$ 2} & \textbf{52}{\scriptsize $\pm$ 2} & \textbf{75}{\scriptsize $\pm$ 2} & \textbf{41}{\scriptsize $\pm$ 2} \\

\bottomrule
\end{tabular}
}

  }
  \caption{
    Success rates on the rearrangement task for our method, \method (highlighted in blue), and baselines.
    Displayed are the average and standard deviations for 3 seeds for M3, Monolithic RL, Skill Transformer, and M3 Oracle (numbers from ~\cite{huang2023skill}), and 3 seeds for the remaining methods with the highest numbers per setting bolded.
    Numbers in the \emph{easy} and \emph{hard} columns are averages over 100 episodes and 200 episodes, respectively. The \emph{All Episodes} column is an average across both the \emph{Easy} and \emph{Hard} episodes.
    \method outperforms all baselines in all settings.
  }
  \label{tab:rearrange}
\end{table*}

\section{Experiments}
\label{sec:experiments} 

\subsection{Object-Rearrangement}
\label{sec:exp-rearrange}

In this section, we compare \method with baselines on the Habitat 2.0 Object Rearrangement task~\cite{szot2021habitat}.
For comparison with baselines, we run our experiments using the setup from~\cite{huang2023skill}.
In this task, a Fetch robot~\cite{fetchrobot} must move an object from a specified start position to a desired goal position in an indoor home environment using only onboard sensing.
The agent has no privileged information like existing maps of the environment, 3D object models, or exact object positions.
The Fetch robot senses the world through a $ 256 \times 256$ depth camera mounted on the robot's head, robot joint positions, gripper state, and base egomotion, giving the relative position of the robot from the start of the episode.
The task is specified by a starting 3D object coordinate for the object to move and a 3D goal coordinate to move the object to. 
Both coordinates are specified relative to the robot's position at the start of the episode.
Only the starting object coordinate is specified, and this information is not updated based on the current object position (e.g. if it is moved). 

The robot interacts with the world via a 7DoF arm, a suction gripper attached to the end of the arm, and a mobile base. 
The episode is successful if the target object is within 15cm of the goal position. The robot has a budget of 1,500 steps to complete the task.

We report performance on the \emph{easy} and \emph{hard} evaluation episodes from~\cite{huang2023skill}.
In easy episodes, both the target object and goal are on an open receptacle. This means the robot does not have to first open a receptacle before accessing the target object or goal. 
Instead, the agent can always execute the same sequence of navigate to object, pick up object, navigate to goal, and place object at goal.
In hard episodes, the object may start in a closed receptacle.
The agent, therefore, needs to use its visual input to perceive if the target object is in a closed receptacle. If so, the robot must then open the receptacle before picking the object. 
The object may start in either the fridge or a cabinet.  

\subsubsection{Training Setup: }
\label{sec:training-setup}
We train \method using 11,791 episodes with a mix of \emph{easy} and \emph{hard} episodes across $ 63$ scenes using the same rearrangement training dataset as in~\cite{habitatrearrangechallenge2022}. These episodes were obtained by sampling an equal number of episodes in the \textit{easy} and \textit{hard} categories, with \textit{easy} episodes being equally sampled across episodes with \textit{open-cabinet}, \textit{open-fridge} and non-articulated episodes. For \textit{hard} episodes, sampling is done uniformly between \textit{closed fridge} and \textit{closed cabinet} episodes. We use auxiliary tasks covering abilities included as a part of the standard rearrangement benchmark used in ~\cite{habitatrearrangechallenge2022}. The auxiliary tasks are:
\begin{itemize}[itemsep=2pt,topsep=0pt,parsep=0pt,partopsep=0pt,parsep=0pt,leftmargin=*]
  \item \emph{Pick}: The agent spawns randomly within the house and must navigate to and pick up the object. 
  \item \emph{Place} the agent is spawned randomly in the house with the object in its gripper. The robot must navigate to and place the object within 15cm of the target 3D location on the receptacle.
  \item \emph{Open Fridge}: The robot spawns randomly within the house and must navigate to the fridge and use its base and arm to open the fridge door.
  \item \emph{Open Cabinet}: The robot spawns randomly in the house and must navigate to the kitchen area and open the cabinet using its base and arm.
  \item \emph{Pick from Fridge: } This task is similar to \emph{Pick} except the agent has to pick up the object from inside an open fridge, which requires careful manipulation to minimize collisions.
\end{itemize}
The auxiliary relevance function is grounded in the oracle task plan~\cite{fikes1971strips} of the episode. This is formulated as an indicator function with $ w_i (s_t) = 1$ if the state $s_{t}$ is relevant to $\mathcal{M}_{i}$ based on the oracle task plan of the episode and $w_i (s_t) = 0$ otherwise.

For example, if at step $ t$ the agent is not holding an object and the object is in an open receptacle, the agent must first pick up the object so $ w_{i_{pick}}(s_t) = 1 $, and $ w_i (s_t) = 0$ for all other skills.
For a complete description of the auxiliary tasks, including their reward functions, along with the precise auxiliary relevance function definition in object rearrangement, see ~\Cref{sec:exp-details}. 
We train methods for 475M steps of environment interactions. We found this to be a sufficient number of environment interactions to ensure sufficient progress on auxiliary tasks for distillation to aid rearrangement performance on both the \textit{easy} and \textit{hard} episodes. For \method, we count the environment interactions in the auxiliary tasks towards this 475M step budget. We train with a learning rate of $ 3e^{-4}$ and linearly decay the learning rate to $ 0$ over the course of learning.

\begin{table*}[t]
  \centering
  \begin{tabular}{lcccccc}
\toprule
& \multicolumn{3}{c}{\textbf{Train} (Seen)} & \multicolumn{3}{c}{\textbf{Eval} (Unseen)} \\
\cmidrule(rl){2-4} \cmidrule(rl){5-7}
Method & \textbf{All Episodes} & \textbf{Easy}  &  \textbf{Hard}  & \textbf{All Episodes}  &  \textbf{Easy}  &  \textbf{Hard}  \\
\midrule
\textbf{All Skills} & 28 & 36 & 19 & 30 & 39 & 21 \\ \textbf{No Pick From Fridge} & 20 & 32 & 7 & 12 & 23 & 2 \\ \textbf{No Open Fridge} & \textbf{41} & \textbf{49} & \textbf{33} & \textbf{48} & \textbf{57} & \textbf{39} \\ \textbf{No Pick} & 0 & 0 & 0 & 0 & 0 & 0 \\
\bottomrule
\end{tabular}

  \caption{
  Robustness to auxiliary task selection. We compare the train and generalization performance of \method with 4 different auxiliary task selections. Results are averages across 100 episodes on a single seed of training. The train and evaluation episodes are in a simplified setting compared to \Cref{tab:rearrange} as described in \Cref{sec:exp-analysis}}.
  \label{tab:ablation}
\end{table*}

\subsubsection{Baselines}
We compare \method to both hierarchical methods governed by a task plan as well as monolithic baselines, which directly learn a pixels-to-actions policy using sensor observations. We compare to relevant baselines from \citet{huang2023skill} and two more baselines that use the auxiliary tasks.
\begin{itemize}[itemsep=2pt,topsep=0pt,parsep=0pt,partopsep=0pt,parsep=0pt,leftmargin=*]
  \item \textbf{Monolithic RL:}  A monolithic neural network is trained to map sensor observations directly to actions trained with end-to-end RL. The policy showed no signs of learning the main task, so we stopped training early after $100M$ steps of training. This outcome is consistent with the monolithic RL baseline in prior works~\cite{berges2023galactic,szot2021habitat}.

    \item \textbf{Galactic:}~\cite{berges2023galactic} An end-to-end RL framework similar to monolithic RL to map sensor observations directly to actions with the policy being trained with a simplified kinematic simulation on over $>1e^{9}$ simulation steps and transferred to dynamic simulation utilized in our setting. 
    
    \item \textbf{M3:}~\cite{gu2022multi} Each of the skill policies navigate, pick, place, and open are separately trained. These skills are then sequenced using a task planner. 

    \item \textbf{M3 (Oracle)}:~\cite{gu2022multi} This uses the same skill training as M3, but uses an oracle task planner.

    \item \textbf{Skill Transformer}~\cite{huang2023skill}: This method uses pre-trained skills to collect successful rearrangement demonstrations and then train on these demonstrations with imitation learning.

    \item \textbf{RL Curriculum} This method is trained in two stages: i) First, we train the policy on the auxiliary tasks for $200M$ steps to ensure the relevant auxiliary tasks have good training performance ii) This is followed by a training on the entire rearrangement task for another $300M$ steps.  

    \item \textbf{\method: No-Distillation} Train \method under a similar setting as described in ~\Cref{sec:training-setup} except with the distillation strength $\lambda = 0.0$. 
\end{itemize}
For more details on the baselines, see \Cref{sec:baseline-details}.

\begin{figure}
    \centering
    \includegraphics[width = 0.8\textwidth]{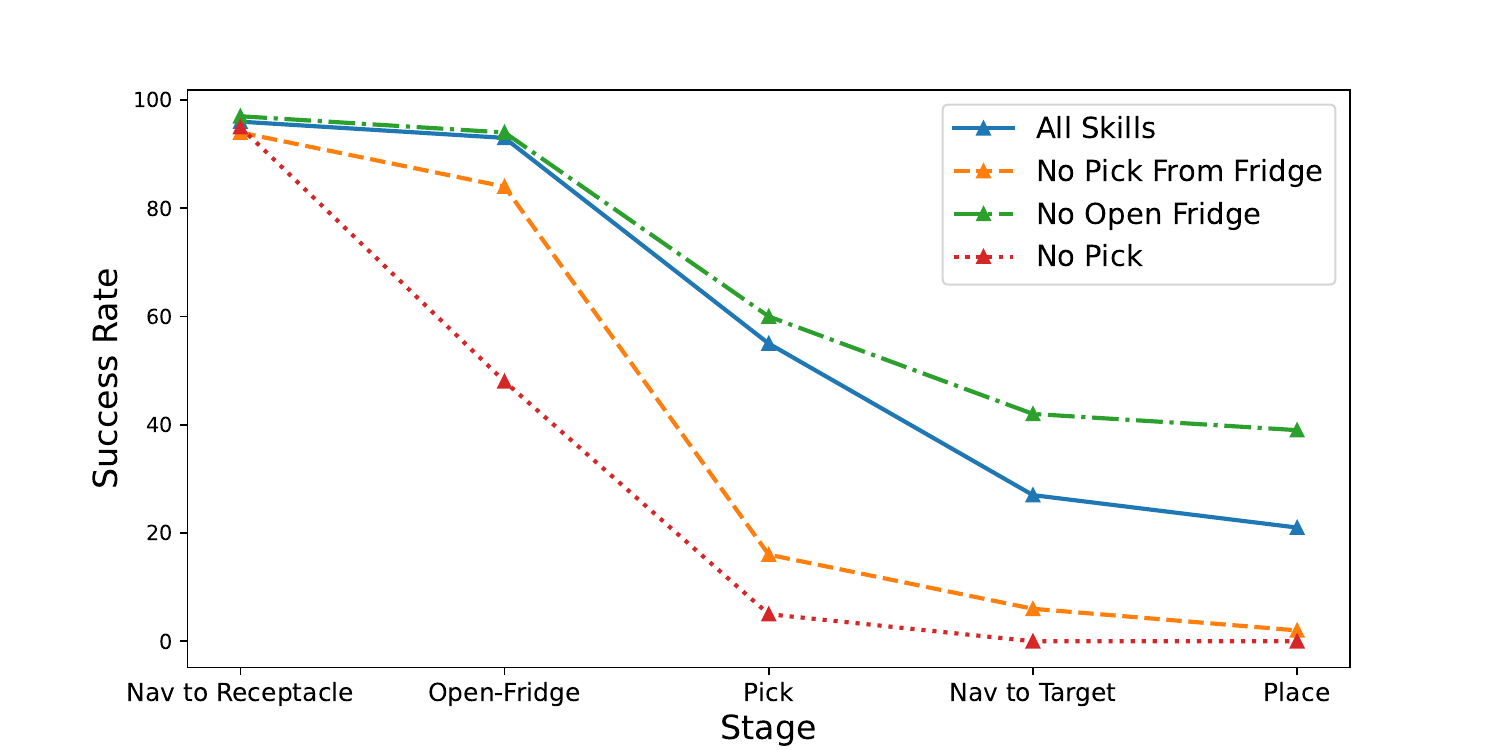}
    \caption{Comparison of skill-robustness to different choices of auxiliary task on the hard distribution. Including both the  \textit{Pick} and \textit{Pick from Fridge} is crucial for successful rearrangement on this distribution. Not utilizing Open-Fridge leads to a boost in rearrangement success. This improvement arises because the open-fridge skill is the easiest of all auxiliary tasks and utilizing it reduces the number of samples for the main task (from the $100$M step budget). We discuss the auxiliary task learning curves in Appendix~\ref{sec:aux-task-learn}}
    \label{fig:stage-robust-plot}
\end{figure}

\subsubsection{Rearrangement Performance: }
In \Cref{tab:rearrange}, we compare \method to baselines in the rearrangement task. On all evaluation settings, \method outperforms all baselines.
\method shows an absolute improvement of $ 22\% $ and $ 25 \% $ on the \emph{easy} and \emph{hard} episodes respectively over the best performing baseline.
Most baselines struggle to achieve any performance on the unseen episodes.
Monolithic RL~\cite{szot2021habitat} achieves no success in any of the evaluation settings, demonstrating the impracticality of learning directly from the dense reward in the full rearrangement task. Our method also outperforms Galactic~\cite{berges2023galactic} despite learning on less than half the number of training samples ($475M$ for \method vs. $>1e^{9}$ for Galactic) on the easy split with $74\%$ success over $37\%$ on easy episodes during training and $75\%$ vs. $26\%$ on evaluation.  Note that Galactic does not report performance on the \textit{hard} distribution.

\begin{figure}[t]
  \centering
   \begin{subfigure}[t]{0.45\columnwidth}
    \includegraphics[width=\linewidth]{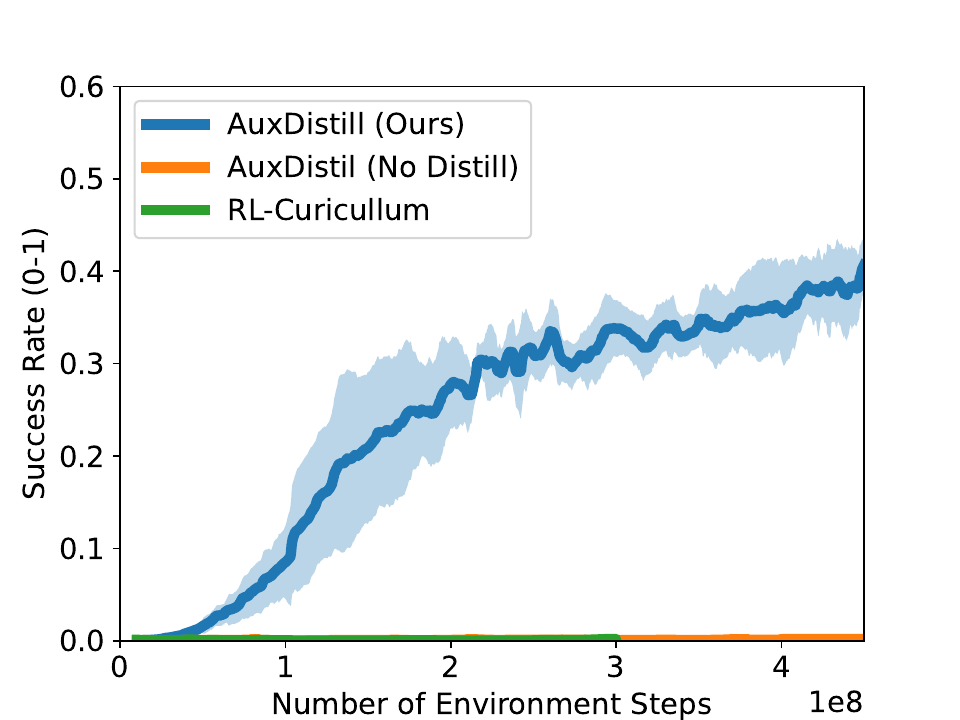}
    \caption{Rearrange learning curve.}
    \label{fig:rearrange-learn}
  \end{subfigure}
   \begin{subfigure}[t]{0.45\columnwidth}
    \includegraphics[width=\linewidth]{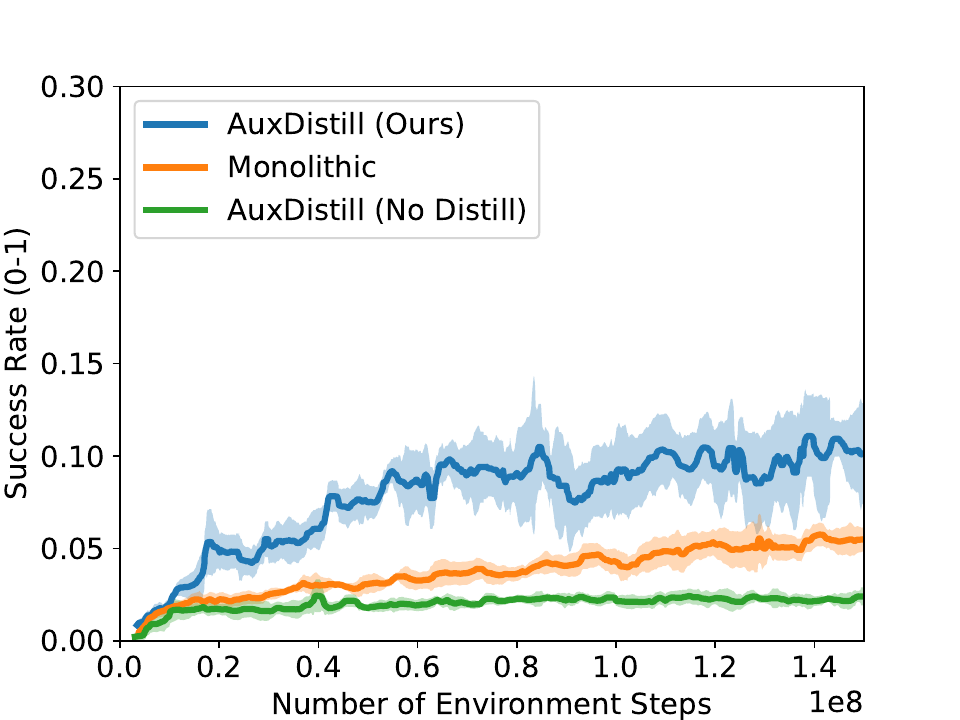}
    \caption{\pick learning.}
    \label{fig:pick-learn}
  \end{subfigure}
  \caption{
    Left: RL training success rates on training episodes on the rearrangement task from ~\Cref{tab:rearrange}. Note that \method (No Distill) and RL-Curriculum are displayed but achieve $0\%$ success throughout training.
    Right: Learning curve on the \pick task of \method utilizing coordinate pick distillation vs. monolithic RL. \method outperforms baselines in both settings.
    Displayed are averages and standard deviation over 3 random seeds.
  }
  \label{fig:hab-overview}
\end{figure}

Like \method, RL Curriculum also utilizes the auxiliary tasks yet finds no success.
This demonstrates the value of concurrent training on the main and auxiliary tasks with the distillation loss in \method.
Likewise, \method (No Distill) achieves no success, illustrating the importance of the distillation loss. 
The learning curve in \Cref{fig:rearrange-learn} as \method can gradually learn to complete the task with more environment interactions.
On the other hand, \curric and \nodist remain at no success regardless of the number of learning samples.

\method also outperforms the strongest baseline, Skill Transformer, by a significant margin with $ 75 \% $ vs. $ 37 \% $ on the unseen easy episodes and $ 41 \% $ vs. $ 16 \% $ on the unseen hard episodes.
This demonstrates the advantages of using online, end-to-end RL, rather than solely training offline with demonstrations. 
As shown in \Cref{fig:rearrange-learn}, \method is able to improve with subsequent environment interactions, yet Skill Transformer is limited by the performance of the expert demonstrations.

\Cref{tab:rearrange} also demonstrates that \method outperforms hierarchical baselines.
M3 is able to achieve $ 53 \% $ success on the unseen \emph{easy} episodes since it utilizes strong pretrained skills.
Even in this setting, \method achieves a higher success of $ 75 \% $. 
However, on the hard episodes, M3 cannot dynamically plan skills and achieves no success in the \emph{hard} unseen episodes. 
\method achieves $ 41 \% $ on this setting because it learns a monolithic policy with RL that combines low-level and high-level decision-making.
\method even outperforms an oracle version of M3 that dynamically plans the skill sequence based on oracle state information ($ 13 \%$ vs. $ 41 \% $ on the unseen hard episodes).

We observe that \method performs $7\%$ worse on the Habitat 2.0 rearrangement training episodes than the evaluation episodes. 
We find this performance gap is due to additional easier episodes in the evaluation distribution where the object is closer to the target location. 
In the hard split, there are $55$ eval episodes and $29$ train episodes where the object is picked up from the fridge and the target location is $<0.5m$ from the object location. On removing these episodes, train performance is higher than evaluation, with $ 32\%$ success on train and $ 31\%$ success on evaluation.

\begin{table*}[t]
  \centering
  \resizebox{0.8 \textwidth}{!}{
    \scalebox{0.1}{
\begin{tabular}{rccc}
\toprule
 Method & {\textbf{Train (Seen)}} & {\textbf{Eval} (Unseen)} \\
\midrule
\textbf{\method} & \textbf{12}{\scriptsize $\pm$ 2} & \textbf{11}{\scriptsize $\pm$ 1} \\ \textbf{Monolithic} & 8{\scriptsize $\pm$ 0} & 9{\scriptsize $\pm$ 1} \\ \textbf{\method (No Distillation)} & 3{\scriptsize $\pm$ 1} & 4{\scriptsize $\pm$ 1} \\ \textbf{RL-Curicullum} & 6{\scriptsize $\pm$ 1} & 8{\scriptsize $\pm$ 0} \\ 
\bottomrule
\end{tabular}
}

  }
  \caption{
    Comparison of our method on the \pick task with auxiliary tasks on three random seeds. \method outperforms all methods, including the monolithic baseline, on the unseen evaluation split. Evaluation is conducted over 1000 seen episodes sampled from the training distribution and 1000 held-out episodes from evaluation.  
  }
  \label{tab:lang_pick}
\end{table*}

\subsection{\pick}
\label{sec:cat-pick}
The merits of \method extend to other challenging embodied tasks. In particular, consider a variant of the pick task described in Section~\ref{sec:exp-rearrange} where a robot has to pick up an object using the object name  passed as a one-hot embedding to the policy. Category pick requires the policy to discern the object to pick up by correlating the RGB observation with the object category passed as a one-hot embedding.  This task is more challenging than geometric pick where the policy has access to the initial coordinate specification of the object to be picked. We include additional details about the task specification in Section~\ref{sec:supp-lang-pick} of our supplementary material. 

We leverage the easier coordinate pick task to aid the learning of the \pick task using \method. More specifically, we consider the task of interest $M_{0}$ to be the \pick task and $M_{1}$ (auxiliary task we would like to distill from) to be the coordinate pick task. Both tasks are trained jointly using \method with the same reward structure with $w_{i}(s_{t}) = 1$ and use $\lambda = 0.5$ for distillation in ~\Cref{eq:main_loss}. For this task, we train on the full rearrange-easy dataset from ~\cite{habitatrearrangechallenge2022} with  50,000 episodes. 

We report the training curves of the \pick task in comparison with the RL Curriculum and No Distillation baseline used in ~\Cref{tab:rearrange}. In addition, we introduce a monolithic baseline for this task, which directly trains \pick without leveraging the coordinate pick task during training. We report the training curves in Figure~\ref{fig:pick-learn}. We evaluate the performance of the trained policy on $1000$ episodes sampled from the training and $1000$ episodes from the validation split. 

In Table~\ref{tab:lang_pick}, we show that \method shows better performance with a success rate of $11\%$ outperforming all baselines that do not leverage distillation during RL training. The closest performing baseline to ours is the monolithic baseline, which achieves a success of $9\%$ on the held-out distribution. \method (No-Distillation) and RL-curriculum perform worse with an evaluation success of $4\%$ and $8\%$, respectively. 
\subsection{\method Analysis}
\label{sec:exp-analysis} 

\subsubsection{Robustness to Auxiliary Task Selection}
A critical consideration for training \method is the selection of auxiliary tasks. 
In this section, we compare four different selections of auxiliary tasks in the object rearrangement task from \Cref{sec:exp-rearrange}. 
We conduct this ablation on a smaller distribution of episodes, which only include the \emph{Easy} episodes and \emph{Hard} episodes where the object starts in the fridge and train \method for 100M steps.  
 
As we conduct this study only on the fridge category of articulated episodes, we modify the auxiliary task selection to be: \textit{Pick}, \textit{Open Fridge}, \textit{Place} and \textit{Pick from Fridge}. Note that the first three auxiliary tasks are the same from \Cref{sec:exp-rearrange}.  The other auxiliary task selections are the same as this original selection, but excluding one of: \textit{Pick}, \textit{Open Fridge}, and \textit{Pick from Fridge}. 

In Table~\ref{tab:ablation}, we report the performance for each skill selection on the easy and hard episodes. We observe that among all the \textit{sub-tasks}, the most important is \textit{Pick}, as excluding it results in $0.0\%$ success. Note that  \textit{Pick} is only relevant ($w_{i} (s_{t}) > 0$) to easy episodes during training i.e removing it should only impact the performance on \textit{easy} distribution of our training setup. However, we find that it affects both the easy and hard episodes as the task is not successfully carried out in either case if removed. In ~\Cref{fig:stage-robust-plot}, we analyze the success of individual stages of rearrangement and notice that not including \textit{Pick} leads to failure earlier on in the task. 

In ~\Cref{tab:aux-skills} we notice that introducing the Open-Fridge skill can worsen rearrangement performance ($30\%$ vs. $48\%$). The reason for this is that the open-fridge skill is the easiest of all auxiliary tasks (see ~\Cref{sec:aux-task-learn}), and utilizing it reduces the number of samples allocated to the main task.  However, certain auxiliary tasks can boost performance by addressing specific bottlenecks in the rearrangement. One such example is the \textit{Pick from Fridge}, which boosts performance  $7\%$ to $19\%$ by addressing the task of picking from an open fridge, which can be challenging due to the difficulty of avoiding collisions with the fridge door. Further, the $4\%$ performance improvement from $32\%$ to $36\%$ on  \textit{easy} distribution can be attributed to the presence of \textit{easy} episodes in our dataset with an open-refrigerator where \textit{Pick from Fridge} helps boost performance. 
 
\subsubsection{Distillation Coefficient Variation}

We analyze the effect of the distillation loss coefficient $ \lambda$ in \method in \Cref{tab:dist-coef}. The distillation loss coefficient controls how the agent balances minimizing the distillation from the auxiliary tasks versus maximizing the reward on the main task. In \Cref{tab:dist-coef}, we notice that using high values of distillation weighting $\lambda = 1.0$  shifts the objective of the rearrangement task from maximizing cumulative reward to distilling from the auxiliary task leading to $1\%$ success rate on rearrangement and similarly, using a very small value of distillation coefficient $\lambda = 0.01$ leads to insufficient leveraging of auxiliary task information with a success rate of $7\%$.  We find $\lambda = 0.1, 0.05$ to be good choices for optimizing task reward and distilling behaviors from the auxiliary tasks. 

\begin{table*}[t]
  \centering
  \resizebox{0.8 \textwidth}{!}{
    \scalebox{0.4}{
\begin{tabular}{rcccccc}
\toprule
& \multicolumn{3}{c}{\textbf{Train} (Seen)} & \multicolumn{3}{c}{\textbf{Eval} (Unseen)} \\
\cmidrule(rl){2-4} \cmidrule(rl){5-7}
Method & \textbf{All Episodes} & \textbf{Easy}  &  \textbf{Hard}  & \textbf{All Episodes}  &  \textbf{Easy}  &  \textbf{Hard}  \\
\midrule
\textbf{$\lambda=0.01$} & 6 & 6 & 5 & 10 & 18 & 2 \\ \textbf{$\lambda=0.05$} & \textbf{30} & \textbf{31} & \textbf{28} & \textbf{36} & 33 & \textbf{38} \\ \textbf{$\lambda=0.1$} & 22 & 26 & 19 & 30 & \textbf{39} & 21 \\ \textbf{$\lambda=0.5$} & 2 & 3 & 0 & 2 & 5 & 0 \\ \textbf{$\lambda=1.0$} & 0 & 1 & 0 & 1 & 2 & 0 \\
\bottomrule
\end{tabular}
}

  }
  \caption{
    Performance of our method for a single seed on varying distillation coefficient during training. Using a large distillation coefficient ($\lambda = 1.0$) makes reward optimization challenging, and too small ($\lambda = 0.01$) results in insufficient auxiliary distillation to succeed on the rearrangement task. The intermediate values $\lambda = 0.05$ and $\lambda = 0.1$ show the best performance on training and evaluation. 
  }
  \label{tab:dist-coef}
\end{table*}

\section{Conclusion}

In this work, we presented \method, a new method for end-to-end RL on complex tasks by leveraging auxiliary tasks.
\method learns in the auxiliary tasks concurrently with the main task through multi-task RL. 
A distillation loss transfers relevant behaviors from the auxiliary tasks to the main task.
We show that \method outperforms a variety of baselines in Habitat object rearrangement by up to 27$\%$.
We also show another application of \method in a category-conditioned manipulation task.
Finally, we analyze \method in different auxiliary task selections and magnitudes of distillation strengths.
Overall, \method presents a new way to tackle compound tasks with RL alone without requiring pre-trained skills or expert demonstrations.

Limitations of \method include the need for the auxiliary tasks and the auxiliary task relevance function.
The auxiliary tasks require knowing behaviors that are relevant and easier to learn than the main task. 
Each auxiliary task requires defining a new start state distribution and reward function. However, designing these can rely on privileged state information which is less restrictive than the requirement of pre-trained skills or expert demonstrations.

{\small
\bibliographystyle{unsrtnat}
\setlength{\bibsep}{0pt}
\bibliography{main}

\begin{thebibliography}{52}
\providecommand{\natexlab}[1]{#1}
\providecommand{\url}[1]{\texttt{#1}}
\expandafter\ifx\csname urlstyle\endcsname\relax
  \providecommand{\doi}[1]{doi: #1}\else
  \providecommand{\doi}{doi: \begingroup \urlstyle{rm}\Url}\fi

\bibitem[Mnih et~al.(2013)Mnih, Kavukcuoglu, Silver, Graves, Antonoglou,
  Wierstra, and Riedmiller]{mnih2013playing}
Volodymyr Mnih, Koray Kavukcuoglu, David Silver, Alex Graves, Ioannis
  Antonoglou, Daan Wierstra, and Martin Riedmiller.
\newblock Playing atari with deep reinforcement learning.
\newblock \emph{arXiv preprint arXiv:1312.5602}, 2013.

\bibitem[Berner et~al.(2019)Berner, Brockman, Chan, Cheung, D{\k{e}}biak,
  Dennison, Farhi, Fischer, Hashme, Hesse, et~al.]{berner2019dota}
Christopher Berner, Greg Brockman, Brooke Chan, Vicki Cheung, Przemys{\l}aw
  D{\k{e}}biak, Christy Dennison, David Farhi, Quirin Fischer, Shariq Hashme,
  Chris Hesse, et~al.
\newblock Dota 2 with large scale deep reinforcement learning.
\newblock \emph{arXiv preprint arXiv:1912.06680}, 2019.

\bibitem[Silver et~al.(2017)Silver, Schrittwieser, Simonyan, Antonoglou, Huang,
  Guez, Hubert, Baker, Lai, Bolton, et~al.]{silver2017mastering}
David Silver, Julian Schrittwieser, Karen Simonyan, Ioannis Antonoglou, Aja
  Huang, Arthur Guez, Thomas Hubert, Lucas Baker, Matthew Lai, Adrian Bolton,
  et~al.
\newblock Mastering the game of go without human knowledge.
\newblock \emph{nature}, 550\penalty0 (7676):\penalty0 354--359, 2017.

\bibitem[Shah et~al.(2016)Shah, Hakkani-T{\"u}r, and Heck]{shah2016interactive}
Pararth Shah, Dilek Hakkani-T{\"u}r, and Larry Heck.
\newblock Interactive reinforcement learning for task-oriented dialogue
  management.
\newblock In \emph{NIPS 2016 Deep Learning for Action and Interaction
  Workshop}, volume~11, 2016.

\bibitem[Liu et~al.(2017)Liu, Tur, Hakkani-Tur, Shah, and Heck]{liu2017end}
Bing Liu, Gokhan Tur, Dilek Hakkani-Tur, Pararth Shah, and Larry Heck.
\newblock End-to-end optimization of task-oriented dialogue model with deep
  reinforcement learning.
\newblock \emph{arXiv preprint arXiv:1711.10712}, 2017.

\bibitem[Nakano et~al.(2021)Nakano, Hilton, Balaji, Wu, Ouyang, Kim, Hesse,
  Jain, Kosaraju, Saunders, et~al.]{nakano2021webgpt}
Reiichiro Nakano, Jacob Hilton, Suchir Balaji, Jeff Wu, Long Ouyang, Christina
  Kim, Christopher Hesse, Shantanu Jain, Vineet Kosaraju, William Saunders,
  et~al.
\newblock Webgpt: Browser-assisted question-answering with human feedback.
\newblock \emph{arXiv preprint arXiv:2112.09332}, 2021.

\bibitem[Achiam et~al.(2023)Achiam, Adler, Agarwal, Ahmad, Akkaya, Aleman,
  Almeida, Altenschmidt, Altman, Anadkat, et~al.]{achiam2023gpt}
Josh Achiam, Steven Adler, Sandhini Agarwal, Lama Ahmad, Ilge Akkaya,
  Florencia~Leoni Aleman, Diogo Almeida, Janko Altenschmidt, Sam Altman,
  Shyamal Anadkat, et~al.
\newblock Gpt-4 technical report.
\newblock \emph{arXiv preprint arXiv:2303.08774}, 2023.

\bibitem[Touvron et~al.(2023)Touvron, Lavril, Izacard, Martinet, Lachaux,
  Lacroix, Rozi{\`e}re, Goyal, Hambro, Azhar, et~al.]{touvron2023llama}
Hugo Touvron, Thibaut Lavril, Gautier Izacard, Xavier Martinet, Marie-Anne
  Lachaux, Timoth{\'e}e Lacroix, Baptiste Rozi{\`e}re, Naman Goyal, Eric
  Hambro, Faisal Azhar, et~al.
\newblock Llama: Open and efficient foundation language models.
\newblock \emph{arXiv preprint arXiv:2302.13971}, 2023.

\bibitem[Akkaya et~al.(2019)Akkaya, Andrychowicz, Chociej, Litwin, McGrew,
  Petron, Paino, Plappert, Powell, Ribas, et~al.]{akkaya2019solving}
Ilge Akkaya, Marcin Andrychowicz, Maciek Chociej, Mateusz Litwin, Bob McGrew,
  Arthur Petron, Alex Paino, Matthias Plappert, Glenn Powell, Raphael Ribas,
  et~al.
\newblock Solving rubik's cube with a robot hand.
\newblock \emph{arXiv preprint arXiv:1910.07113}, 2019.

\bibitem[Qi et~al.(2023)Qi, Kumar, Calandra, Ma, and Malik]{qi2023hand}
Haozhi Qi, Ashish Kumar, Roberto Calandra, Yi~Ma, and Jitendra Malik.
\newblock In-hand object rotation via rapid motor adaptation.
\newblock In \emph{Conference on Robot Learning}, pages 1722--1732. PMLR, 2023.

\bibitem[Kalashnikov et~al.(2018)Kalashnikov, Irpan, Pastor, Ibarz, Herzog,
  Jang, Quillen, Holly, Kalakrishnan, Vanhoucke, and Levine]{qt-opt}
Dmitry Kalashnikov, Alex Irpan, Peter Pastor, Julian Ibarz, Alexander Herzog,
  Eric Jang, Deirdre Quillen, Ethan Holly, Mrinal Kalakrishnan, Vincent
  Vanhoucke, and Sergey Levine.
\newblock Scalable deep reinforcement learning for vision-based robotic
  manipulation.
\newblock In \emph{2nd Annual Conference on Robot Learning, CoRL 2018,
  Z{\"{u}}rich, Switzerland, 29-31 October 2018, Proceedings}, volume~87 of
  \emph{Proceedings of Machine Learning Research}, pages 651--673. {PMLR},
  2018.
\newblock URL \url{http://proceedings.mlr.press/v87/kalashnikov18a.html}.

\bibitem[Batra et~al.(2020)Batra, Chang, Chernova, Davison, Deng, Koltun,
  Levine, Malik, Mordatch, Mottaghi, et~al.]{batra2020rearrangement}
Dhruv Batra, Angel~X Chang, Sonia Chernova, Andrew~J Davison, Jia Deng, Vladlen
  Koltun, Sergey Levine, Jitendra Malik, Igor Mordatch, Roozbeh Mottaghi,
  et~al.
\newblock Rearrangement: A challenge for embodied ai.
\newblock \emph{arXiv preprint arXiv:2011.01975}, 2020.

\bibitem[Harutyunyan et~al.(2019)Harutyunyan, Dabney, Mesnard, Gheshlaghi~Azar,
  Piot, Heess, van Hasselt, Wayne, Singh, Precup,
  et~al.]{harutyunyan2019hindsight}
Anna Harutyunyan, Will Dabney, Thomas Mesnard, Mohammad Gheshlaghi~Azar, Bilal
  Piot, Nicolas Heess, Hado~P van Hasselt, Gregory Wayne, Satinder Singh, Doina
  Precup, et~al.
\newblock Hindsight credit assignment.
\newblock \emph{Advances in neural information processing systems}, 32, 2019.

\bibitem[Ni et~al.(2024)Ni, Ma, Eysenbach, and Bacon]{ni2024transformers}
Tianwei Ni, Michel Ma, Benjamin Eysenbach, and Pierre-Luc Bacon.
\newblock When do transformers shine in rl? decoupling memory from credit
  assignment.
\newblock \emph{Advances in Neural Information Processing Systems}, 36, 2024.

\bibitem[Berges et~al.(2023)Berges, Szot, Chaplot, Gokaslan, Mottaghi, Batra,
  and Undersander]{berges2023galactic}
Vincent-Pierre Berges, Andrew Szot, Devendra~Singh Chaplot, Aaron Gokaslan,
  Roozbeh Mottaghi, Dhruv Batra, and Eric Undersander.
\newblock Galactic: Scaling end-to-end reinforcement learning for rearrangement
  at 100k steps-per-second.
\newblock In \emph{Proceedings of the IEEE/CVF Conference on Computer Vision
  and Pattern Recognition}, pages 13767--13777, 2023.

\bibitem[Huang et~al.(2023)Huang, Batra, Rai, and Szot]{huang2023skill}
Xiaoyu Huang, Dhruv Batra, Akshara Rai, and Andrew Szot.
\newblock Skill transformer: A monolithic policy for mobile manipulation.
\newblock \emph{arXiv preprint arXiv:2308.09873}, 2023.

\bibitem[Gu et~al.(2022)Gu, Chaplot, Su, and Malik]{gu2022multi}
Jiayuan Gu, Devendra~Singh Chaplot, Hao Su, and Jitendra Malik.
\newblock Multi-skill mobile manipulation for object rearrangement.
\newblock \emph{arXiv preprint arXiv:2209.02778}, 2022.

\bibitem[Narvekar et~al.(2020)Narvekar, Peng, Leonetti, Sinapov, Taylor, and
  Stone]{narvekar2020curriculum}
Sanmit Narvekar, Bei Peng, Matteo Leonetti, Jivko Sinapov, Matthew~E Taylor,
  and Peter Stone.
\newblock Curriculum learning for reinforcement learning domains: A framework
  and survey.
\newblock \emph{The Journal of Machine Learning Research}, 21\penalty0
  (1):\penalty0 7382--7431, 2020.

\bibitem[Dennis et~al.(2020)Dennis, Jaques, Vinitsky, Bayen, Russell, Critch,
  and Levine]{dennis2020emergent}
Michael Dennis, Natasha Jaques, Eugene Vinitsky, Alexandre Bayen, Stuart
  Russell, Andrew Critch, and Sergey Levine.
\newblock Emergent complexity and zero-shot transfer via unsupervised
  environment design.
\newblock \emph{Advances in neural information processing systems},
  33:\penalty0 13049--13061, 2020.

\bibitem[Azad et~al.(2023)Azad, Gur, Emhoff, Alexis, Faust, Abbeel, and
  Stoica]{azad2023clutr}
Abdus~Salam Azad, Izzeddin Gur, Jasper Emhoff, Nathaniel Alexis, Aleksandra
  Faust, Pieter Abbeel, and Ion Stoica.
\newblock Clutr: Curriculum learning via unsupervised task representation
  learning.
\newblock In \emph{International Conference on Machine Learning}, pages
  1361--1395. PMLR, 2023.

\bibitem[Fang et~al.(2022)Fang, Migimatsu, Mandlekar, Fei-Fei, and
  Bohg]{fang2022active}
Kuan Fang, Toki Migimatsu, Ajay Mandlekar, Li~Fei-Fei, and Jeannette Bohg.
\newblock Active task randomization: Learning visuomotor skills for sequential
  manipulation by proposing feasible and novel tasks.
\newblock \emph{arXiv preprint arXiv:2211.06134}, 2022.

\bibitem[Szot et~al.(2021)Szot, Clegg, Undersander, Wijmans, Zhao, Turner,
  Maestre, Mukadam, Chaplot, Maksymets, et~al.]{szot2021habitat}
Andrew Szot, Alexander Clegg, Eric Undersander, Erik Wijmans, Yili Zhao, John
  Turner, Noah Maestre, Mustafa Mukadam, Devendra~Singh Chaplot, Oleksandr
  Maksymets, et~al.
\newblock Habitat 2.0: Training home assistants to rearrange their habitat.
\newblock \emph{Advances in Neural Information Processing Systems}, 34, 2021.

\bibitem[Sutton et~al.(1999)Sutton, Precup, and Singh]{sutton1999between}
Richard~S Sutton, Doina Precup, and Satinder Singh.
\newblock Between mdps and semi-mdps: A framework for temporal abstraction in
  reinforcement learning.
\newblock \emph{Artificial intelligence}, 112\penalty0 (1-2):\penalty0
  181--211, 1999.

\bibitem[Bacon et~al.(2017)Bacon, Harb, and Precup]{bacon2017option}
Pierre-Luc Bacon, Jean Harb, and Doina Precup.
\newblock The option-critic architecture.
\newblock In \emph{Proceedings of the AAAI Conference on Artificial
  Intelligence}, volume~31, 2017.

\bibitem[Zhang and Whiteson(2019)]{zhang2019dac}
Shangtong Zhang and Shimon Whiteson.
\newblock Dac: The double actor-critic architecture for learning options.
\newblock \emph{Advances in Neural Information Processing Systems}, 32, 2019.

\bibitem[Xia et~al.(2020)Xia, Li, Mart{\'\i}n-Mart{\'\i}n, Litany, Toshev, and
  Savarese]{xia2020relmogen}
Fei Xia, Chengshu Li, Roberto Mart{\'\i}n-Mart{\'\i}n, Or~Litany, Alexander
  Toshev, and Silvio Savarese.
\newblock Relmogen: Leveraging motion generation in reinforcement learning for
  mobile manipulation.
\newblock \emph{arXiv preprint arXiv:2008.07792}, 2020.

\bibitem[Karkus et~al.(2020)Karkus, Mirza, Guez, Jaegle, Lillicrap, Buesing,
  Heess, and Weber]{karkus2020beyond}
Peter Karkus, Mehdi Mirza, Arthur Guez, Andrew Jaegle, Timothy Lillicrap, Lars
  Buesing, Nicolas Heess, and Theophane Weber.
\newblock Beyond tabula-rasa: a modular reinforcement learning approach for
  physically embedded 3d sokoban.
\newblock \emph{arXiv preprint arXiv:2010.01298}, 2020.

\bibitem[Dalal et~al.(2021)Dalal, Pathak, and
  Salakhutdinov]{dalal2021accelerating}
Murtaza Dalal, Deepak Pathak, and Russ~R Salakhutdinov.
\newblock Accelerating robotic reinforcement learning via parameterized action
  primitives.
\newblock \emph{Advances in Neural Information Processing Systems},
  34:\penalty0 21847--21859, 2021.

\bibitem[Hafner et~al.(2022)Hafner, Lee, Fischer, and Abbeel]{hafner2022deep}
Danijar Hafner, Kuang-Huei Lee, Ian Fischer, and Pieter Abbeel.
\newblock Deep hierarchical planning from pixels.
\newblock \emph{Advances in Neural Information Processing Systems},
  35:\penalty0 26091--26104, 2022.

\bibitem[Vezzani et~al.(2022)Vezzani, Tirumala, Wulfmeier, Rao, Abdolmaleki,
  Moran, Haarnoja, Humplik, Hafner, Neunert, et~al.]{vezzani2022skills}
Giulia Vezzani, Dhruva Tirumala, Markus Wulfmeier, Dushyant Rao, Abbas
  Abdolmaleki, Ben Moran, Tuomas Haarnoja, Jan Humplik, Roland Hafner, Michael
  Neunert, et~al.
\newblock Skills: Adaptive skill sequencing for efficient temporally-extended
  exploration.
\newblock \emph{arXiv preprint arXiv:2211.13743}, 2022.

\bibitem[Chen et~al.(2023)Chen, Wang, Fei-Fei, and Liu]{chen2023sequential}
Yuanpei Chen, Chen Wang, Li~Fei-Fei, and C~Karen Liu.
\newblock Sequential dexterity: Chaining dexterous policies for long-horizon
  manipulation.
\newblock \emph{arXiv preprint arXiv:2309.00987}, 2023.

\bibitem[Mishra et~al.(2023)Mishra, Xue, Chen, and Xu]{mishra2023generative}
Utkarsh~Aashu Mishra, Shangjie Xue, Yongxin Chen, and Danfei Xu.
\newblock Generative skill chaining: Long-horizon skill planning with diffusion
  models.
\newblock In \emph{Conference on Robot Learning}, pages 2905--2925. PMLR, 2023.

\bibitem[Lee et~al.(2021)Lee, Lim, Anandkumar, and Zhu]{lee2021adversarial}
Youngwoon Lee, Joseph~J Lim, Anima Anandkumar, and Yuke Zhu.
\newblock Adversarial skill chaining for long-horizon robot manipulation via
  terminal state regularization.
\newblock \emph{arXiv preprint arXiv:2111.07999}, 2021.

\bibitem[Schulman et~al.(2017)Schulman, Wolski, Dhariwal, Radford, and
  Klimov]{schulman2017proximal}
John Schulman, Filip Wolski, Prafulla Dhariwal, Alec Radford, and Oleg Klimov.
\newblock Proximal policy optimization algorithms.
\newblock \emph{arXiv preprint arXiv:1707.06347}, 2017.

\bibitem[Rudin et~al.(2022)Rudin, Hoeller, Reist, and
  Hutter]{rudin2022learning}
Nikita Rudin, David Hoeller, Philipp Reist, and Marco Hutter.
\newblock Learning to walk in minutes using massively parallel deep
  reinforcement learning.
\newblock In \emph{Conference on Robot Learning}, pages 91--100. PMLR, 2022.

\bibitem[Agarwal et~al.(2023)Agarwal, Kumar, Malik, and
  Pathak]{agarwal2023legged}
Ananye Agarwal, Ashish Kumar, Jitendra Malik, and Deepak Pathak.
\newblock Legged locomotion in challenging terrains using egocentric vision.
\newblock In \emph{Conference on Robot Learning}, pages 403--415. PMLR, 2023.

\bibitem[Fu et~al.(2023)Fu, Cheng, and Pathak]{fu2023deep}
Zipeng Fu, Xuxin Cheng, and Deepak Pathak.
\newblock Deep whole-body control: learning a unified policy for manipulation
  and locomotion.
\newblock In \emph{Conference on Robot Learning}, pages 138--149. PMLR, 2023.

\bibitem[{Kumar} et~al.(2021){Kumar}, {Fu}, {Pathak}, and
  {Malik}]{kumar2021rma}
Ashish {Kumar}, Zipeng {Fu}, Deepak {Pathak}, and Jitendra {Malik}.
\newblock Rma: Rapid motor adaptation for legged robots.
\newblock \emph{RSS}, 2021.

\bibitem[Radosavovic et~al.(2023)Radosavovic, Xiao, Zhang, Darrell, Malik, and
  Sreenath]{radosavovic2023learning}
Ilija Radosavovic, Tete Xiao, Bike Zhang, Trevor Darrell, Jitendra Malik, and
  Koushil Sreenath.
\newblock Learning humanoid locomotion with transformers.
\newblock \emph{arXiv preprint arXiv:2303.03381}, 2023.

\bibitem[Katara et~al.(2023)Katara, Xian, and Fragkiadaki]{katara2023gen2sim}
Pushkal Katara, Zhou Xian, and Katerina Fragkiadaki.
\newblock Gen2sim: Scaling up robot learning in simulation with generative
  models.
\newblock \emph{arXiv preprint arXiv:2310.18308}, 2023.

\bibitem[Ye et~al.(2021{\natexlab{a}})Ye, Batra, Wijmans, and
  Das]{ye2021auxiliary}
Joel Ye, Dhruv Batra, Erik Wijmans, and Abhishek Das.
\newblock Auxiliary tasks speed up learning point goal navigation.
\newblock In \emph{Conference on Robot Learning}, pages 498--516. PMLR,
  2021{\natexlab{a}}.

\bibitem[Ye et~al.(2021{\natexlab{b}})Ye, Batra, Das, and
  Wijmans]{ye2021auxiliary2}
Joel Ye, Dhruv Batra, Abhishek Das, and Erik Wijmans.
\newblock Auxiliary tasks and exploration enable objectgoal navigation.
\newblock In \emph{Proceedings of the IEEE/CVF International Conference on
  Computer Vision}, pages 16117--16126, 2021{\natexlab{b}}.

\bibitem[Gregor et~al.(2019)Gregor, Jimenez~Rezende, Besse, Wu, Merzic, and
  van~den Oord]{gregor2019shaping}
Karol Gregor, Danilo Jimenez~Rezende, Frederic Besse, Yan Wu, Hamza Merzic, and
  Aaron van~den Oord.
\newblock Shaping belief states with generative environment models for rl.
\newblock \emph{Advances in Neural Information Processing Systems}, 32, 2019.

\bibitem[Kuo et~al.(2023)Kuo, Ma, Hoffman, and Kira]{kuo2023structure}
Chia-Wen Kuo, Chih-Yao Ma, Judy Hoffman, and Zsolt Kira.
\newblock Structure-encoding auxiliary tasks for improved visual representation
  in vision-and-language navigation.
\newblock In \emph{Proceedings of the IEEE/CVF Winter Conference on
  Applications of Computer Vision}, pages 1104--1113, 2023.

\bibitem[Jia et~al.(2022)Jia, Li, Ling, Liu, Wu, and Su]{jia2022improving}
Zhiwei Jia, Xuanlin Li, Zhan Ling, Shuang Liu, Yiran Wu, and Hao Su.
\newblock Improving policy optimization with generalist-specialist learning.
\newblock In \emph{International Conference on Machine Learning}, pages
  10104--10119. PMLR, 2022.

\bibitem[Baker et~al.(2019)Baker, Kanitscheider, Markov, Wu, Powell, McGrew,
  and Mordatch]{baker2019emergent}
Bowen Baker, Ingmar Kanitscheider, Todor Markov, Yi~Wu, Glenn Powell, Bob
  McGrew, and Igor Mordatch.
\newblock Emergent tool use from multi-agent autocurricula.
\newblock \emph{arXiv preprint arXiv:1909.07528}, 2019.

\bibitem[He et~al.(2016)He, Zhang, Ren, and Sun]{he2016deep}
Kaiming He, Xiangyu Zhang, Shaoqing Ren, and Jian Sun.
\newblock Deep residual learning for image recognition.
\newblock \emph{CVPR}, 2016.

\bibitem[Hessel et~al.(2019)Hessel, Soyer, Espeholt, Czarnecki, Schmitt, and
  Van~Hasselt]{hessel2019multi}
Matteo Hessel, Hubert Soyer, Lasse Espeholt, Wojciech Czarnecki, Simon Schmitt,
  and Hado Van~Hasselt.
\newblock Multi-task deep reinforcement learning with popart.
\newblock In \emph{Proceedings of the AAAI Conference on Artificial
  Intelligence}, volume~33, pages 3796--3803, 2019.

\bibitem[robotics(2020)]{fetchrobot}
Fetch robotics.
\newblock Fetch.
\newblock \url{http://fetchrobotics.com/}, 2020.

\bibitem[Szot et~al.(2022)Szot, Yadav, Clegg, Berges, Gokaslan, Chang, Savva,
  Kira, and Batra]{habitatrearrangechallenge2022}
Andrew Szot, Karmesh Yadav, Alex Clegg, Vincent-Pierre Berges, Aaron Gokaslan,
  Angel Chang, Manolis Savva, Zsolt Kira, and Dhruv Batra.
\newblock Habitat rearrangement challenge 2022.
\newblock \url{https://aihabitat.org/challenge/2022_rearrange}, 2022.

\bibitem[Fikes and Nilsson(1971)]{fikes1971strips}
Richard~E Fikes and Nils~J Nilsson.
\newblock Strips: A new approach to the application of theorem proving to
  problem solving.
\newblock \emph{Artificial intelligence}, 2\penalty0 (3-4):\penalty0 189--208,
  1971.

\bibitem[Wijmans et~al.(2019)Wijmans, Kadian, Morcos, Lee, Essa, Parikh, Savva,
  and Batra]{wijmans2019dd}
Erik Wijmans, Abhishek Kadian, Ari Morcos, Stefan Lee, Irfan Essa, Devi Parikh,
  Manolis Savva, and Dhruv Batra.
\newblock Dd-ppo: Learning near-perfect pointgoal navigators from 2.5 billion
  frames.
\newblock \emph{arXiv preprint arXiv:1911.00357}, 2019.

\end{thebibliography}
}

\newpage
\appendix

\section{Additional Method Details}
\label{sec:method-details} 
\subsection{Pseudocode}
  
\begin{algorithm}[H]
    \caption{The workflow for training \method from a randomly initialized policy}
    \label {algo:method}
    Initialize policy $\pi_{\theta}$ \\
    Initialize state-space, observation space and action space $\{\mathcal{S}, \mathcal{O}, \mathcal{A}\}$ \\
    Define relevance $w_{i}: \mathcal{S} \rightarrow \{0, 1\} \hspace{1mm} \forall \hspace{1mm} i \in \{1, 2, \cdots N\}$\\
    Initialize distillation coefficient, normalization parameters $\lambda, \beta = 3e^{-4}$ \\
    \For{$epoch \leftarrow 1$ \KwTo train-epochs}
    {
        // bold face represents a vector.  \\
        Collect a batch of $B$ samples  $\{\mathbf{r}_{i}\, \mathbf{o}_{i}, \mathbf{s}_{i}. \mathbf{a}_{i} \}_{i=0}^{N}$ by executing $\{\pi_{\theta}^{T_{i}}\}_{i=0}^{N}$ \\
        Normalize returns per task: $\{\mathbf{r}^{n}_{i}\}_{i=0}^{N} = PopArt(\{\mathbf{r}_{i}\}_{i=0}^{N}, \beta)$ \\
        Obtain $\{\pi_\theta^{T_i} (\mathbf{o}_{0})\}_{i=1}^{N}$ by evaluating $\mathbf{o}_{0}$ with $\{T_{i}\}_{i=1}^{N}$  \\
        \vspace{2mm} 
        // Compute losses \\
        Compute RL losses: $L_{RL}$ using $\{\mathbf{r}^{n}_{i}, \mathbf{o}_{i}, \mathbf{s}_{i}, \mathbf{a}_{i}\}_{i=0}^{N}$ \\
        Compute distill loss $L_{d}$:  $\frac{1}{N} \sum_{i=1}^{N} \sum_{t=0}^{B} w_{i}({s}^{t}_{0})   \kl{\pi_\theta^{T_0} ({o}^{t}_0)}{\pi_\theta^{T_i} ({o}^{t}_0)}$ \\
        Update $\pi_\theta$ using $L_{distill} = L_{RL}$ + $\lambda L_{d}$
    
    }
\end{algorithm}

In this section, we describe the workflow to implement \method, including data collection for all of the tasks in our training mix, computing the distillation loss, and policy optimization. We describe our algorithm in~\Cref{algo:method}

We initialize a policy $\pi_{\theta}$ along with the state space $\mathcal{S}$, observation space $\mathcal{O}$ and action space $\mathcal{A}$.  The relevance function $\{w_{i}(\mathcal{S})\}_{i=1}^{N} \rightarrow \{0,1\}$ is defined as a mapping between the robot state to a real-valued relevance used for distillation. The distillation weights are assigned to $\lambda=0.1$ and $\beta=3e^{-4}$. For each epoch of our training cycle, we collect a batch of data by executing our policy on each of the tasks and collect a batch of returns $\{\mathbf{r}_{i}\}_{i=1}^{N}$, observations $\{\mathbf{o}_{i}\}_{i=1}^{N}$, states $\{\mathbf{s}_{i}\}_{i=1}^{N}$ and actions $\{\mathbf{a}_{i}\}_{i=1}^{N}$. The returns are then normalized using PopArt with parameter $\beta$ as $\{\mathbf{r}^{n}_{i}\}_{i=1}^{N}$. The observations of the main task are evaluated under each of the tasks $T_{i}$ to obtain $\{\pi^{T_{i}}_{\theta} (\mathbf{o}_{0})\}_{i=1}^{N}$. Note that here $\mathbf{s}_{i}$ represents the complete state information at a given time step, whereas $\mathbf{o}_{i}$ represents the visual observation available to the agent.

As described in ~\Cref{sec:method}, our policy optimizes a weighted combination of two objectives. The RL-loss required for a regular PPO update given by $L_{RL}$ losses computed using normalized returns, observations, states and actions $\{\mathbf{r}_{i}^{n}, \mathbf{o}_{i}^{n}, \mathbf{s}_{i}^{n}, \mathbf{a}_{i}^{n}\}$. The distillation loss is computed for each time step, i.e. $\{t = 1, 2, \cdots B\}$ as the KL-divergence between $\kl{\pi_\theta^{T_0} ({o}^{t}_0)}{\pi_\theta^{T_i} ({o}^{t}_0)}$ weighted by the task relevance ${w}_{i}(s_{0}^{t})$. The total loss for policy optimization is computed using the weighting parameter $\lambda$ as $L_{\text{distill}} = L_{RL} + \lambda L_{d}$. 

\subsection{Additional Implementation Details}
In this section, we further detail network architecture and implementation details used in the training of \method. The agent captures visual observations using a $256 \times 256$ depth sensor, which is encoded by a ResNet50~\cite{he2016deep} architecture. The visual features are concatenated with the proprioceptive and goal sensor observations and passed onto the LSTM backbone network. The LSTM architecture has $2$ hidden layers with $128$ hidden units per layer, which generates a state representation of the environment. The LSTM output features are regressed to a multi-task value-head using a 2-layer critic network after concatenating with the task indicator $T$. The features generated by the output layer of the LSTM are used to regress to $\{\mu, \sigma\} \in \mathbb{R}^{10}$. The actions are then sampled from a Gaussian distribution defined by $\{\mu, \sigma\}$, i.e., $ a_{t} \sim \mathcal{N}(\mu, \sigma)$. Overall, our network has $\approx 13M$ trainable parameters. 

The policy is updated using DDPPO~\cite{wijmans2019dd}, which collects data from $24$ environment workers parallelized across $8$ GPUs across $6$ tasks (5 auxiliary tasks and the main task). The policy is updated after collecting  $128$ steps of experience in each worker with a pre-emption threshold of $0.25$. The policy is trained for $450M$ steps collected across all auxiliary tasks. The starting learning rate of training is  $3e^{-4}$ with linear learning rate decay to $ 0$ over $500M$ steps. Before each policy update, we normalize the returns per-task using Pop-Art with $\beta=3.0e^{-4}$ (see ~\cite{hessel2019multi} for more details). 

We utilize PPO with a value loss coefficient of $0.5$ and an entropy coefficient of $0.001$ to incentivize exploration. Considering the longer horizon of the rearrangement task, we set the discount rate $\gamma = 0.999$.   The distillation loss is computed by evaluating the observations of the main task under all task indicators $\{T^{i}\}_{i=1}^{N}$ and computing the weighted KL-divergence using the auxiliary task relevance function. In our experiments, we determine the relevance function using the episode's current robot state and oracle task plan. (see~\Cref{tab:aux-skills} for more details). This is used in conjunction with the RL-losses with a weighting factor of $\lambda=0.1$.

\section{Additional Experiment Details}
\label{sec:exp-details} 

\subsection{Additional Task Details}
We train our method on a setup similar to the Habitat-Rearrangement challenge~\cite{habitatrearrangechallenge2022}, which involves a robot spawned in a home indoor environment with the goal of manipulating objects without having access to privileged maps or other oracle information and solely operating using ego-centric perception. Each episode is specified by the 3D coordinate location of the object and goal at the beginning of the episode. As this dataset is imbalanced across easy and hard episodes, we sub-sample  $11,791$  from the entire dataset of $50,000$ episodes in the rearrange dataset. The episodes are obtained by selecting an equal percentage of episodes with the object spawned inside a closed cabinet, closed fridge, or open receptacles. To carry out the task, we use a Fetch robot with a mobile base and a 7-DOF arm with a gripper. We provide the following inputs to the policy: 

\begin{itemize}[itemsep=2pt,topsep=0pt,parsep=0pt,partopsep=0pt,parsep=0pt,leftmargin=*]
   \item \textbf{Depth Camera:} a $256 \times 256$ depth camera attached to the head of the robot.
    \item \textbf{Coordinate Sensors:}
The euclidean distance between the 3D coordinate of the robot end-effector and the object to be picked up and the location where the object is to be placed. 
    \item \textbf{Holding sensor:} indicating whether the robot is grasping any object. 
    \item \textbf{Relative Resting Position Sensor:} highlighting the Euclidean distance of the 3D coordinate position of the end-effector to the resting position. 
    \item \textbf{Joint sensor:} indicating the 7-DOF joint position of the robot arm.
    \item \textbf{Task Indicator:} encoded as a one-hot vector indicating which of the tasks is currently being executed. 
\end{itemize}
The agent can interact with the environment for a maximum of $1500$ steps. Further, we enforce a step threshold for each task stage based on the auxiliary tasks defined in ~\Cref{sec:supp-aux-task}. During training, we utilize an instantaneous force threshold of $30$kN, but do not apply the force threshold during evaluation as consistent with~\cite{huang2023skill}.

\subsubsection{Auxiliary Task Definitions}
\label{sec:supp-aux-task}
In this section, we describe the auxiliary task definitions for each of the tasks utilized in our training and ablation experiments. 
\\
\\
\textbf{Pick}: This task involves the agent being spawned randomly in the house at least $3m$ from the object of interest without the object in hand. The task is considered successful if the agent is successfully able to navigate to and pick up the object by calling a grip action when it is $0.15m$ from the object of interest and rearrange its arm to $0.15m$ of resting position. The horizon length of this task is $700$ steps. The reward function for this task is represented as: 
\begin{align*}
  R(s_t) = 10 \mathbb {I}_{success} + 2 \Delta^{o}_{arm} \mathbb{I}_{!holding} + 2 \Delta^{r}_{arm} \mathbb{I}_{holding} + 2 \mathbb{I}_{pick}
\end{align*} 
Here, $\mathbb{I}_{pick}$ represents the condition of the pick skill successfully picking up the object, and  $\mathbb{I}_{success}$  represents the agent being able to pick up the object successfully and rearrange its arm to the resting position. $2 \Delta^{r}_{arm} \mathbb{I}_{holding}$ and  $\Delta^{o}_{arm}$ represents the Euclidean distance of the robot of the arm to the object and $\Delta^{r}_{arm}$ represents the deviation from the resting position. 
\\
\\
\textbf{Place:} This task involves the agent being spawned randomly in the house at least $3m$ from the object without the object in hand. The agent has to navigate to the target receptacle, place the object within $0.15m$ of the goal location, and rearrange its arm to its resting position. The horizon length of this task is $700$ steps.  
\\
\begin{align*}
  R(s_t) = 10 \mathbb {I}_{success} + 2 \Delta^{t}_{arm} \mathbb{I}_{holding} + 2 \Delta^{r}_{arm} \mathbb{I}_{!holding} + 5 \mathbb{I}_{place}
\end{align*} 

Here, $\mathbb{I}_{success}$, $\mathbb{I}_{place}$ represent a sparse reward for successful task completion and placing the object, respectively. $\Delta^{t}_{arm}$ arm represents the per-time step deviation of the robot arm to the target location when the agent is holding the object and $\Delta^{r}_{arm}$ represents the deviation of the robot arm towards resting position after the object has been placed successfully. 

\textbf{Open-Cabinet:} This skill involves the robot being spawned randomly in the house, with the task of navigating to the cabinet and opening the drawer by calling the grasp action within $0.15m$ of the drawer handle marker. The drawer is then opened to a joint position of $0.45$. Further, the agent must successfully rearrange its arm to its resting position. The task horizon length for this task is $600$ steps. The reward structure for this task is given by, 
\begin{align*}
  R(s_t) = 10 \mathbb {I}_{success} +  \Delta^{m}_{arm} \mathbb{I}_{!open} + 10 \Delta^{r}_{arm} \mathbb{I}_{open} + 5 \mathbb{I}_{open} + 5 \mathbb{I}_{grasp}
\end{align*} 
Here, $\mathbb{I}_{success}$ is an indicator for successful opening followed by arm rearrangement,  $\mathbb{I}_{open}$ is the indicator for the drawer being successfully opened, and $\mathbb{I}_{grasp}$ is an indicator for the drawer handle being successfully grasped.   $\Delta^{m}_{arm}, \Delta^{r}_{arm}$ are used to encode the dense time-step reward based on the change in arm position to the target marker location and the resting position, respectively. 

\textbf{Open Fridge:} This task involves the robot being spawned randomly in the house, with the task of navigating to the fridge in the scene successfully grasping the fridge handle marker by calling the grasp action at $0.15m$ from the fridge door handle. The fridge door must be opened to a joint position of $1.22$, and the arm must be rearranged to its resting position. The task horizon length for this task $600$ steps. 
The per-time-step reward for this is modeled as: 
\begin{align*}
  R(s_t) = 10 \mathbb {I}_{success} +  \Delta^{m}_{arm} \mathbb{I}_{!open} +  \Delta^{r}_{arm} \mathbb{I}_{open} + 5 \mathbb{I}_{open} + 5 \mathbb{I}_{grasp}
\end{align*} 
Here, $\mathbb{I}_{success}$, $\mathbb{I}_{open}$ and $\mathbb{I}_{grasp}$ are similar to the ones defined for the Open-Cabinet skill.  

\textbf{Pick from Fridge:} This task is similar in structure to the \textbf{Pick} skill except that the data distribution involves picking up an object has to be picked up from an open refrigerator with the agent being spawned $<2m$ from the target object.  
\\
\\
Which of these auxiliary tasks are utilized for distillation is determined by whether the agent's current state is relevant to the state of the agent in the rearrangement task. We show a table of each of the relevance of each task in ~\Cref{tab:aux-skills}. 

\begin{table*}[t]
  \centering
  \resizebox{0.8\textwidth}{!}{
    \begin{tabular}{c|c|c|c}
    \hline
     & \multicolumn{3}{|c}{\textbf{Pre-Conditions for Distillation}} \\
    \hline
  Auxillary Task & {\footnotesize Object Receptacle} & {\footnotesize Did Pick Object?} & {\footnotesize Is Receptacle Open?} \\
    \hline
    \textbf{\footnotesize Pick} & {\footnotesize \textit{Open}} & {\footnotesize $\times$} & {\footnotesize \checkmark} \\
    \hline
    \textbf{\footnotesize Place} & {\footnotesize \textit{Open,Fridge,Cabinet}} & {\footnotesize \checkmark} & {\footnotesize \checkmark, $\times$} \\
    \hline
    \textbf{\footnotesize Open-Fridge} & {\footnotesize \textit{Fridge}} & {\footnotesize $\times$} & {\footnotesize $\times$} \\
    \hline
    \textbf{\footnotesize Open-Cabinet} & {\footnotesize \textit{Cabinet}}  & {\footnotesize $\times$} & {\footnotesize $\times$} \\
    \hline 
    \textbf{\footnotesize Pick from Fridge} & {\footnotesize \textit{Fridge}}  & {\footnotesize $\times$} &  {\footnotesize \checkmark} \\
    \hline
\end{tabular}

  }
  \caption{
    A table representing the relevance of each of the auxiliary tasks based on the stage of the task the robot is in.\textbf{ Did Pick Object?} represents the success of picking up the correct object and \textbf{Is Receptacle Open? } represents whether the robot has successfully opened the receptacle once during the episode. The Object-Receptacle encodes oracle information about the category of episodes we're operating on. The open fridge and open cabinet tasks represent cases when the object is in an open receptacle.
  }
  \label{tab:aux-skills}
\end{table*}

\begin{figure}
    \centering
    \begin{subfigure}[b]{0.3\textwidth}
        \includegraphics[width=\linewidth]{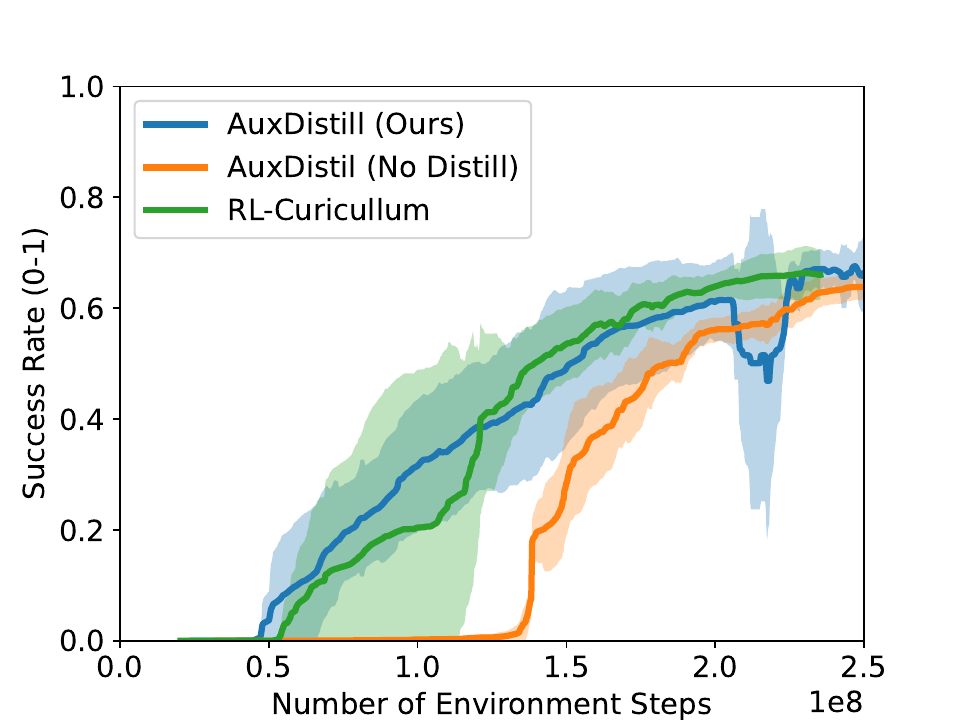}
        \caption{Pick }
        \label{fig:aux-pick}
    \end{subfigure}
    \begin{subfigure}[b]{0.3\textwidth}
        \includegraphics[width=\linewidth]{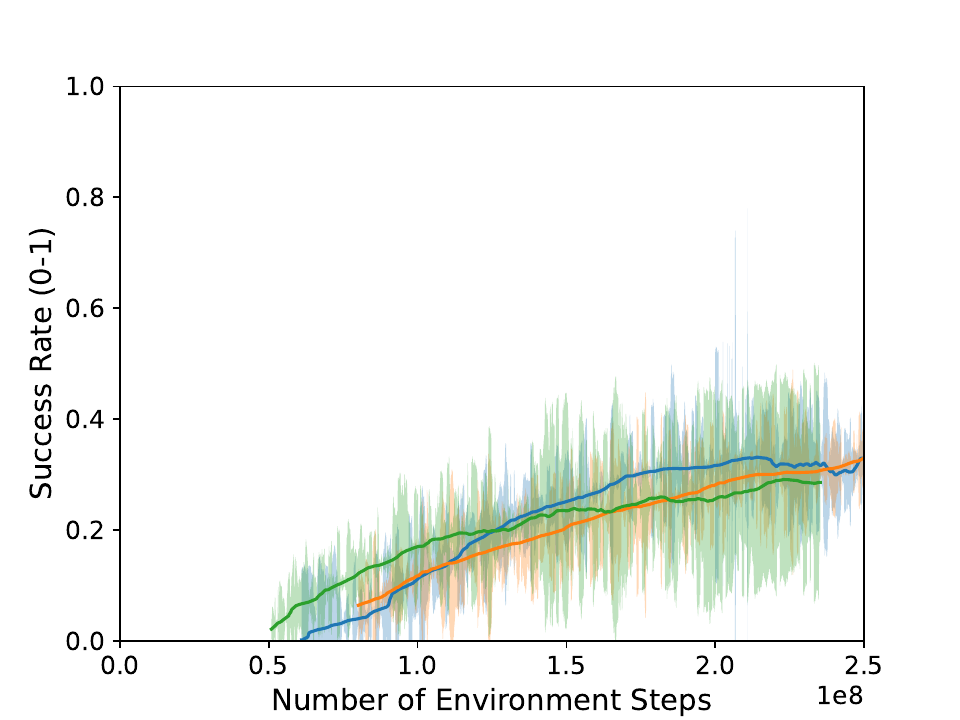}
        \caption{Place}
        \label{fig:aux-place}
    \end{subfigure}
    \begin{subfigure}[b]{0.3\textwidth}
        \includegraphics[width=\linewidth]{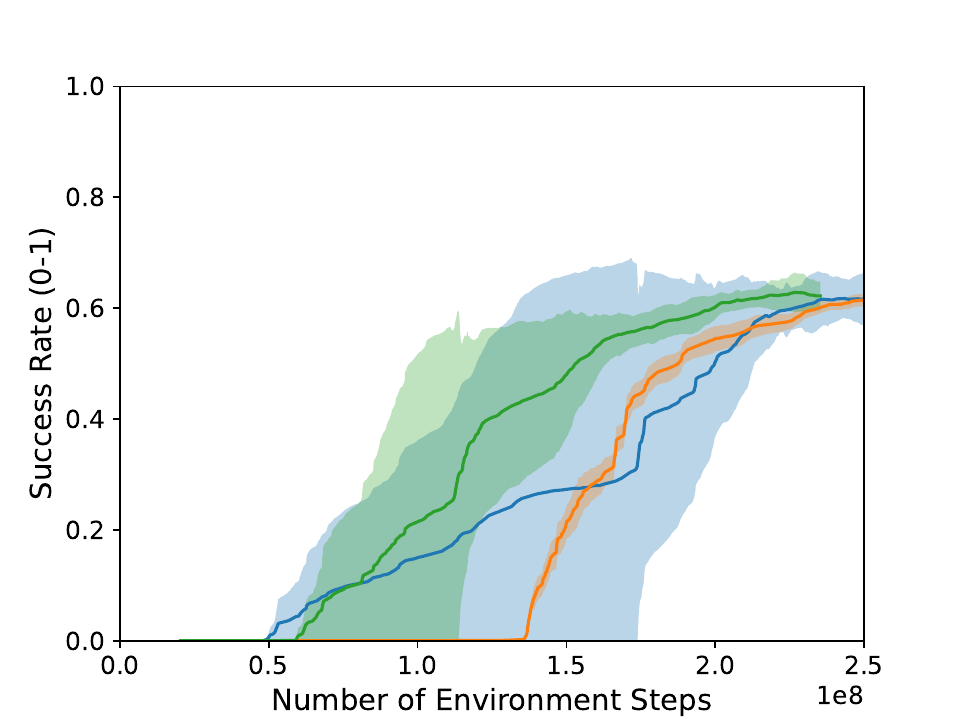}
        \caption{Open-Fridge-Pick}
        \label{fig:aux-of-pick}
    \end{subfigure}
    \\
    \begin{subfigure}[b]{0.3\textwidth}
        \includegraphics[width=\linewidth]{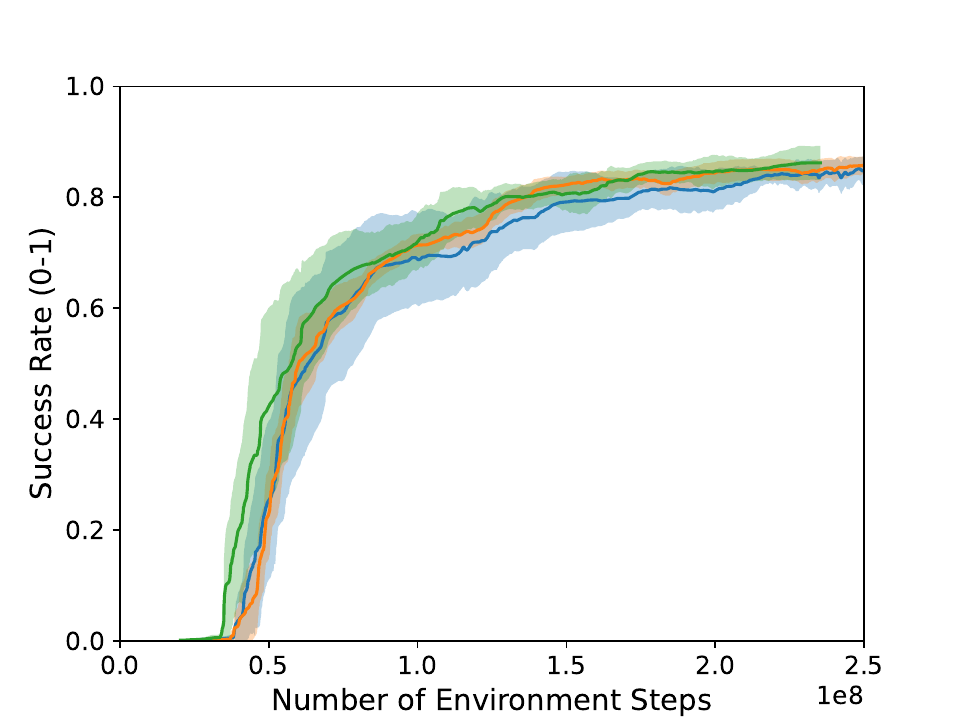}
        \caption{Open-Cabinet}
        \label{fig:sub4}
    \end{subfigure}
    \begin{subfigure}[b]{0.3\textwidth}
        \includegraphics[width=\linewidth]{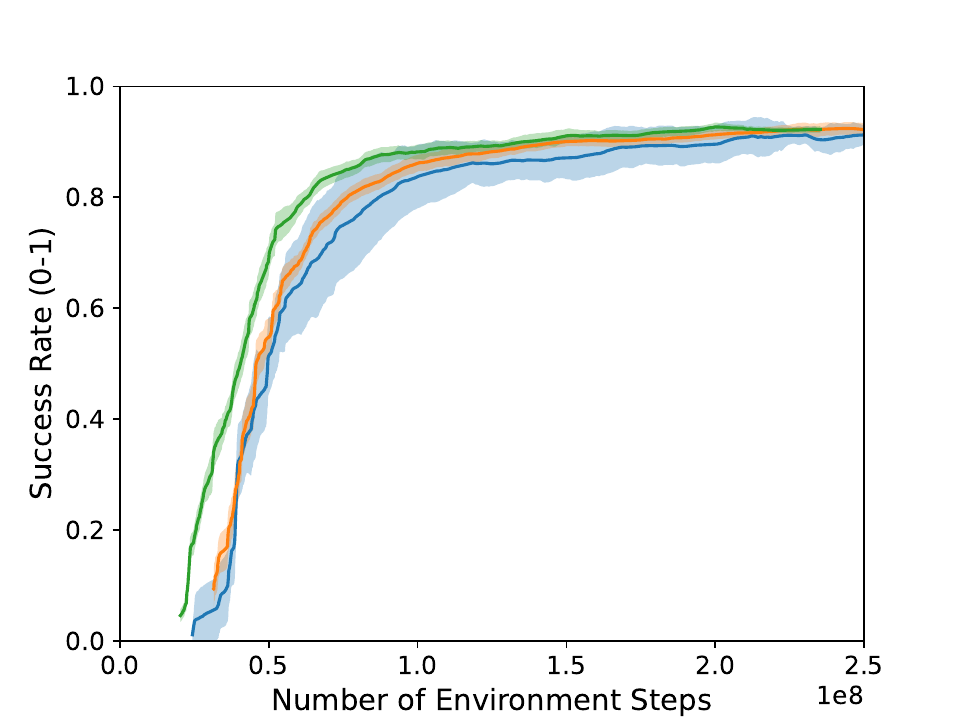}
        \caption{Open-Fridge}
        \label{fig:sub5}
    \end{subfigure}
    \caption{Success curves of the individual skills for the main experiment reported in ~\Cref{tab:rearrange}. The Open-Fridge Pick and Place skill shows high variance across seeds for the main method. RL-Curiculum shows higher variance on the Pick skill. Results are reported up to $250M$ steps number of training steps to show a comparison with all baselines (RL-Curicullum trains sub-skills only for the first stage) }
    \label{fig:aux-skill-learning-curves}
\end{figure}

\subsection{Additional Baseline Details}
\label{sec:baseline-details} 

\subsubsection{M3 $\&$ M3-Oracle} This baseline sequences mobile manipulation policies, which include Navigation, Pick, Place, Open-Fridge, and Open-Cabinet from ~\cite{gu2022multi} each of which are trained for 100M steps using PPO~\cite{schulman2017proximal}. M3 (Oracle) is an oracle version of M3 that has access to privileged information about the oracle sequence of skills based on whether the object begins inside a closed or open receptacle at the start of the episode. Each policy is similar to the ones reported in ~\cite{huang2023skill} with training for  $100M$ steps.
\subsubsection{RL-Curriculum} This baseline captures a 2-stage variant of our method involving the training of all the auxiliary tasks $\{\mathcal{M}_{i}\}_{i=1}^{N}$ at once followed by learning on the main task $\mathcal{M}_0$. The intuition behind this strategy is to learn a policy capable of performing simpler auxiliary tasks, which can be leveraged to learn the challenging main task of interest in the second stage. We allocate a budget of $500M$ steps of agent experience, which is divided into two stages - i) $200M$ steps across all auxiliary tasks followed by ii) $300M$ steps to learn the main task. In the first stage, we encode each of the tasks using a separate identifier $T$. During the second stage, we pass in a task identifier capturing the main rearrangement task unseen during the first stage. We begin each stage with a learning rate of $3e^{-4}$ with linear learning rate decay over $300M$ steps. 

\subsection{Auxillary Task Learning}
\label{sec:aux-task-learn}
In ~\Cref{fig:aux-skill-learning-curves}, we show the results of auxiliary task learning for the main results reported in ~\Cref{tab:rearrange}. We report up to $250M$ steps of learning to demonstrate the results of all baselines, including stage 1 of the curriculum baseline.

Among all the skills, the open and close cabinet skills are the easiest to learn and show the lowest variance during training. The order of difficulty for the other skills is followed by \textit{Pick} < \textit{Pick from Fridge} <\textit{Place}. The difficulty arising in the Placing skill is due to the distinct starting state distribution with the robot being spawned with the object in hand, which none of the other skills have.

\subsection{\pick}
\label{sec:supp-lang-pick}

As described in ~\Cref{sec:cat-pick}, the \pick task demonstrates the merits of \method on a challenging observation space of the object category as opposed to its 3D coordinate location. We encode the object category as a one-hot sensor across a total of $20$ objects. Further, the agent receives an RGB-sensor observation as opposed to the depth sensor used in our rearrangement experiments. Further, the robot is spawned $<2m$ to the target receptacle. All other RL-training parameters are similar to coordinate pick, including the reward structure, episode horizon length, and success condition, which are similar to the ones described in ~\ref{sec:supp-aux-task}.

We train \pick with $16$ environment workers with $8$ environments for each coordinate pick and category-pick parallelized across 8 GPUs. We train for $140M$ steps collected across both language-pick and coordinate pick with a starting learning rate of $2e^{-4}$ with a linear learning rate decay across $200M$ steps of training. We compare this with the monolithic baseline implemented with all $16$ environments devoted to \pick. For training, we utilize the standard rearrange-easy dataset with $50,000$ episodes and evaluate $1,000$ episodes on the standard validation split. 

\begin{figure}
    \centering
    \begin{subfigure}[b]{0.45\textwidth}
        \includegraphics[width=\linewidth]{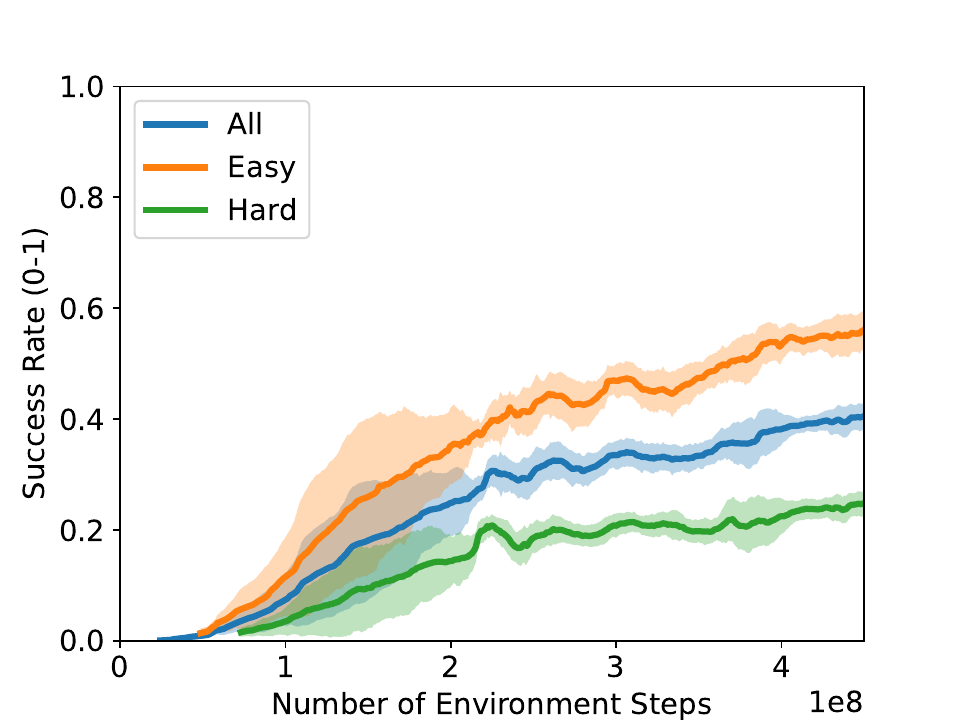}
        \caption{Main Task Learning}
        \label{fig:cur-main}
    \end{subfigure}
    \begin{subfigure}[b]{0.45\textwidth}
        \includegraphics[width=\linewidth]{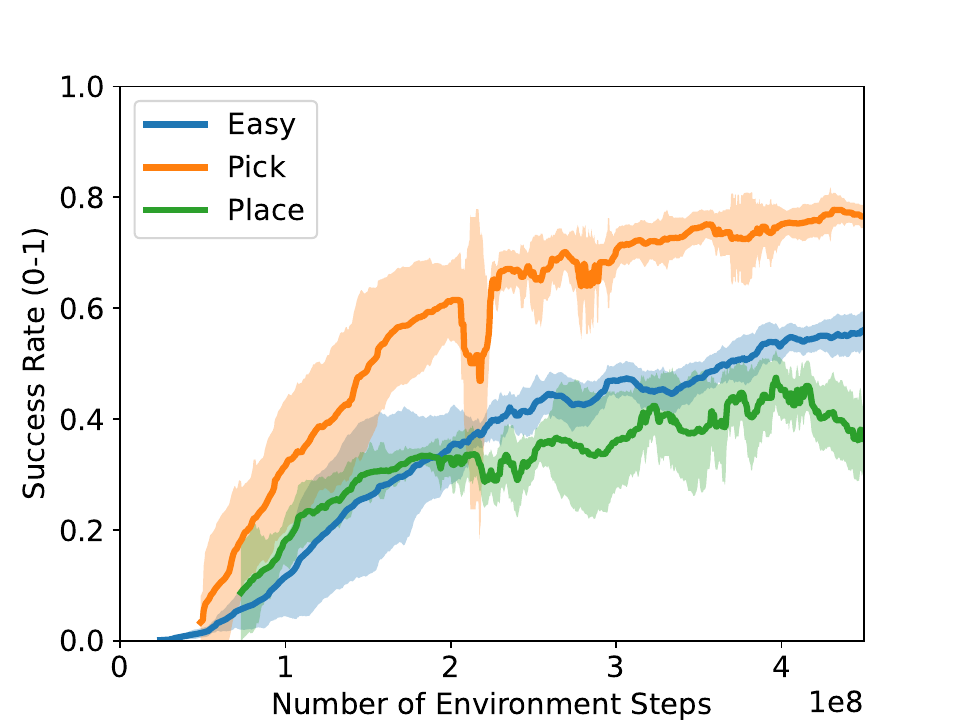}
        \caption{Aux-Task with Easy Main Task}
        \label{fig:cur-easy}
    \end{subfigure}
    \caption{
    Analyzing the curriculum of behaviors that emerges while training \method. On the left we compare the learning of the easier tasks followed by the harder tasks. On the right, we show a comparison of the main task ($\mathcal{M}_{0}$) learning with the relevant auxiliary tasks. On the left, the easier task learns first, followed by the harder task; on the right, the easier distribution improves only after learning the relevant auxiliary skills, i.e., \textit{Pick} and \textit{Place} begin learning.  }
    \label{fig:curicullum}
\end{figure}

\section{Emergent Curriculum During Training}
\vspace{-2mm}
\label{sec:exp-curicullum}
In this section, we discuss how training \method results in a curriculum. Note that our \method does not enforce this curriculum explicitly; this arises as a consequence of the multi-task RL training regime that optimizes to maximize cumulative return. 

In ~\Cref{fig:curicullum}, we show two such curricula; ~\Cref{fig:cur-main}, which shows that the \textit{easy} distribution is learned first during training, followed by the hard distribution. 
Recall that in~\Cref{tab:ablation} we show that rearrangement fails to learn on hard episodes without including the easier \textit{Pick} auxiliary task during training. Building on this observation, in ~\Cref{fig:cur-easy}, we show another curriculum between the auxiliary task and the \textit{easy} main task. The improvement of tasks is in the order \textit{Pick} <    \textit{Place} <  \textit{Rearrange-Easy}. Note that this trend, however, does not hold throughout training; the success rate for the place skill saturates sooner at about $300M$ steps. This difference could be due to the stricter success conditions requiring arm rearrangement for places that are not required by the main easy task.

\end{document}